%% file: main.tex
\documentclass[10pt,twocolumn,letterpaper]{article}

\usepackage[pagenumbers]{cvpr}

\input{preamble}

\definecolor{cvprblue}{rgb}{0.21,0.49,0.74}
\usepackage[pagebackref,breaklinks,colorlinks,allcolors=cvprblue]{hyperref}

\DeclareRobustCommand{\suppref}[1]{Supp.~\cref{#1}}

\def\confName{CVPR}
\def\confYear{2027}

\title{\mbox{SeasonRainBench: Global Seasonal Precipitation}\\
Forecasting from Simulated Worlds to the Real Earth}

\author{
Qiuyan Sun$^{1}$\quad
Mengxuan Chen$^{1}$\quad
Haoyuan Liang$^{1}$\quad
Guowen Li$^{1}$\quad
Zhen Huang$^{1}$\quad
Jianhao Lin$^{1}$\\
Juepeng Zheng$^{2,*}$\quad
Haohuan Fu$^{1,*}$\\[2pt]
{\normalsize $^{1}$Tsinghua University\qquad $^{2}$Sun Yat-sen University\qquad $^{*}$Corresponding authors}
}

\begin{document}
\maketitle
\input{sec/0_abstract}
\input{sec/1_intro}
\input{sec/2_related}
\input{sec/3_benchmark}
\input{sec/4_method}
\input{sec/5_experiments}
\input{sec/6_conclusion}

\section*{LLM Usage Statement}
Large language models were used only to polish the language and formatting of this paper. All research ideas, methods, experiments, analyses and conclusions are the authors' own, and the authors reviewed every edit and take full responsibility for the content.

{
    \small
    \bibliographystyle{ieeenat_fullname}
    \bibliography{main}
}

\input{sec/X_suppl}

\end{document}

%% file: sec/0_abstract.tex
\begin{abstract}
Accurate seasonal precipitation forecasts help societies prepare for drought, flooding, and crop failure months in advance.
While data-driven methods have made rapid progress in weather and subseasonal forecasting, operational seasonal precipitation prediction still relies on numerical ensemble systems.
These systems are computationally expensive and exhibit biases and drift that increase with lead time, motivating more efficient data-driven alternatives.
Yet this transition faces three linked challenges: comparing data-driven forecasts with operational ones requires (1) a global benchmark that does not exist, scoring them requires (2) metrics that capture operational priorities beyond anomaly correlation (ACC) and RMSE, and training them must (3) rely on a single, short observational record.
To address the first two challenges, we introduce \BenchName, the first such benchmark, covering monthly precipitation forecasts 1--5 months ahead.
It pairs global $1^\circ$ satellite precipitation with oceanic and atmospheric predictors, includes seven operational centers, and provides 310,944 simulated months from 52 climate models for training.
Beyond standard scores, we introduce two operationally motivated metrics for anomaly strength and monthly rain-belt position.
To address the third challenge, we develop \MODEL, which learns from simulations what the record lacks and transfers it to Earth through three modules: SimForced aligns the climate background, SimMemory transfers a history encoder pretrained on simulations, and SimAnalog supplements Earth labels with simulated analog outcomes.
Extensive experiments show that \MODEL{} improves ACC, CRPS, anomaly strength, and rain-belt position by 12.0\%, 3.6\%, 2.9\%, and 0.3\% over the strongest operational or data-driven model, opening a path from simulated worlds to Earth system prediction.
\looseness=-1
\end{abstract}

%% file: sec/1_intro.tex
\section{Introduction}
\label{sec:intro}

\begin{figure}[t]
  \centering
  \includegraphics[width=\linewidth]{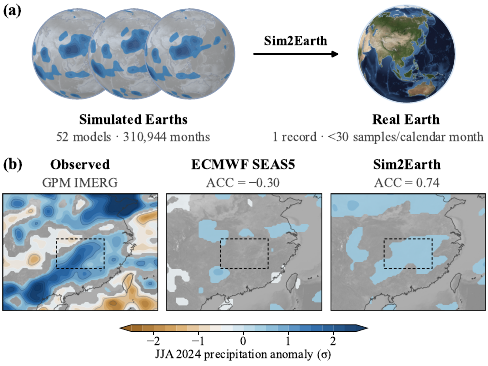}
  \caption{(a) Many simulated records versus one short observed record. (b) Dongting Lake flood, June--August 2024, forecast in May; anomalies below $0.25\sigma$ are unshaded, and ACC is the pattern correlation in the dashed box.}
  \label{fig:teaser}
\end{figure}

\begin{modeloverview}
  \centering
  \includegraphics[width=\textwidth]{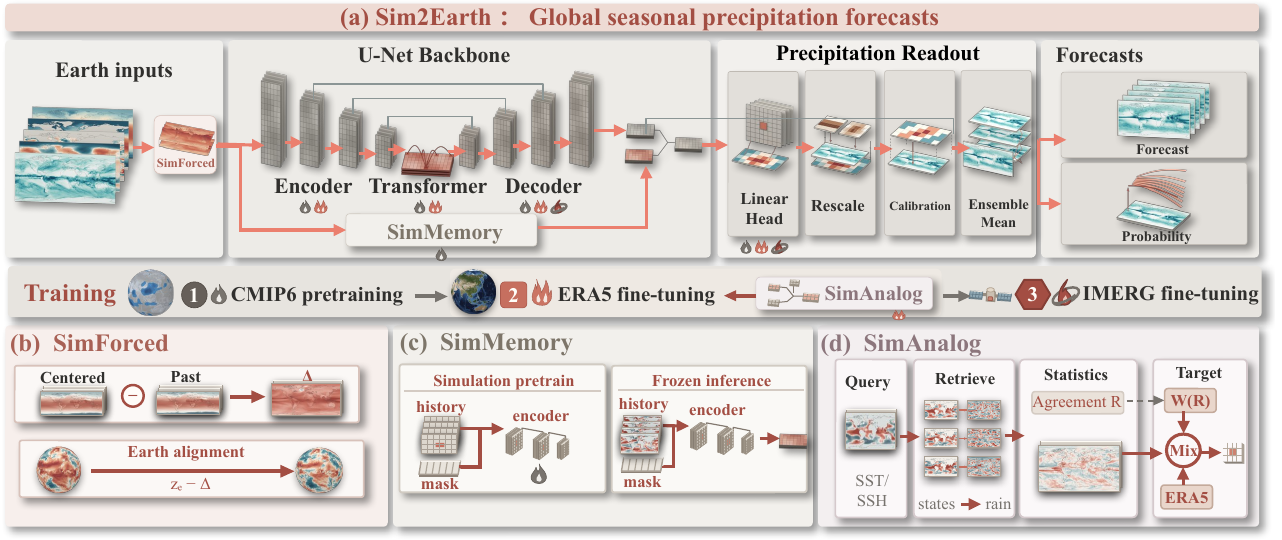}
  \caption{\textbf{\MODEL{} at $1^\circ$ resolution.} (a) Forecast path and three training stages. The probability head maps the mean of calibrated point forecasts to 51 quantile members per lead. (b)--(d) SimForced background alignment, SimMemory seasonal memory, and SimAnalog training-only supervision.}
  \label{fig:arch}
\end{modeloverview}

\begin{benchtable}
  \caption{Comparison with other benchmarks. Among these benchmarks, only \BenchName{} combines global seasonal horizons, simulation training data, satellite verification, and multiple operational baselines.}
  \label{tab:bench}
  \centering
  \begin{SRFitTable}{\textwidth}
\SRTableStyle[2.8pt]
\begin{tabular}{l|ccc|cc|cc}
    \toprule
    \multirow{2}{*}{\textbf{Benchmark}} & \multicolumn{3}{c|}{\textbf{Forecast setting}} & \multicolumn{2}{c|}{\textbf{Data}} & \multicolumn{2}{c}{\textbf{Evaluation}} \\
    \cmidrule(lr){2-4}\cmidrule(lr){5-6}\cmidrule(lr){7-8}
    & Domain & Horizon & Grid & Main sources & Simulation data & Operational baselines & Operational metrics \\
    \midrule
    WeatherBench~2~\cite{rasp2024weatherbench2} & Global & 1--14 days & $1.5^\circ$$^*$ & ERA5 / IFS & --- & 1 center & SEEPS \\
    RainBench~\cite{dewitt2021rainbench} & Global & 1--5 days & $5.625^\circ$ & ERA5 / IMERG & SimSat imagery & --- & --- \\
    SubseasonalClimateUSA~\cite{mouatadid2023subseasonalclimateusa} & Contiguous US & Weeks 3--4 / 5--6 & $1^\circ$ / $1.5^\circ$ & CPC / NCEP / OISST & --- & 2 centers & --- \\
    ChaosBench~\cite{nathaniel2024chaosbench} & Global & 1--44 days & $1.5^\circ$ & ERA5 / ERA5-Land / ORAS5 & --- & 4 centers & --- \\
    SeasonBench-EA~\cite{chen2025seasonbench} & East Asia & 0--5 months$^\dagger$ & $0.25^\circ$ / $1^\circ$ & ERA5 / C3S & --- & 2 centers & EASM hindcast \\
    ClimateSet~\cite{kaltenborn2023climateset} & Global & Emission scenarios & $144\!\times\!96$ & Input4MIPs / CMIP6 & CMIP6: 36 models & --- & --- \\
    \midrule
    \textbf{\BenchName{}} & \textbf{Global} & \textbf{1--5 months} & $\mathbf{1^\circ}$ & \textbf{ERA5 / ERSST / GODAS / IMERG} & \textbf{CMIP6: 52 models} & \textbf{7 centers} & \textbf{AGSS / RBLE} \\
    \bottomrule
  \end{tabular}
\end{SRFitTable}
  \par\smallskip
  \begin{minipage}{\textwidth}\scriptsize
  $^*$Evaluation grid; other entries give data grids (ClimateSet: longitude $\times$ latitude). $^\dagger$Leads 1--6 in our convention, where lead 0 is the issue month. Sources are abbreviated; ours are detailed in \cref{sec:bench}.
  \end{minipage}
\end{benchtable}

Seasonal precipitation forecasting predicts global monthly precipitation one to five months ahead from the observed ocean and atmosphere, specifying the location, strength, and month-to-month evolution of its anomalies (departures from normal) to support agriculture, water management, energy planning, and flood preparedness~\cite{chen2025seasonbench,hirabayashi2013global,zscheischler2018future,raymond2020understanding}.
In July 2024, for example, the longest heavy-rain spell in Hunan since 1961 swelled Dongting Lake until a dike burst, flooding 48\,km$^2$ and forcing over 7,000 residents to evacuate (\cref{fig:teaser}b)~\cite{zhang2024dongting}.
To prepare for such events, operational centers such as ECMWF issue these forecasts monthly~\cite{johnson2019seas5}, and the China Meteorological Administration releases an annual flood-season outlook of where the main rain belt will lie~\cite{zhao2026climate}.
These systems rely on coupled numerical models~\cite{johnson2019seas5,bauer2015quiet}.
However, they remain computationally expensive, drift with lead time, and underuse growing observational and climate-simulation archives~\cite{bauer2015quiet,huffman2020imerg,eyring2016cmip6}.

Data-driven models offer a way to exploit these archives.
Recent models, including Pangu-Weather~\cite{bi2023accurate}, GraphCast~\cite{lam2023learning}, GenCast~\cite{price2025probabilistic}, and FuXi-S2S~\cite{chen2024fuxis2s}, have achieved competitive performance against numerical systems at medium-range and subseasonal horizons.
Data-driven studies of seasonal precipitation, by contrast, remain scarce~\cite{gibson2021training,pan2022improving,palma2026datadriven,pinheiro2025telnet}.
Extending these models to global monthly precipitation one to five months ahead raises three linked challenges:
\looseness=-1

\ding{182}~\textbf{\textit{Comparison lacks a global benchmark.}} Existing benchmarks target medium-range, subseasonal, or regional seasonal forecasting, leaving no unified protocol for comparing global data-driven precipitation forecasts with operational systems~\cite{rasp2024weatherbench2,nathaniel2024chaosbench,chen2025seasonbench}.

\ding{183}~\textbf{\textit{Scoring misses operational priorities.}} Anomaly correlation (ACC) and RMSE summarize association and squared error, but do not directly report the strength of precipitation anomalies or the monthly location of the rain belt, the two quantities that flood-season outlooks state~\cite{chen2025seasonbench,liu2021precursory}.

\ding{184}~\textbf{\textit{Training relies on a single, short record.}} Satellite precipitation offers fewer than thirty samples per calendar month, each the only observed outcome of its ocean state. Climate simulations instead provide many long simulated trajectories (\cref{fig:teaser}a)~\cite{huffman2020imerg,eyring2016cmip6}.

To address these challenges, we present a unified framework that couples a global benchmark with a simulation-to-Earth forecasting model.
The benchmark, \BenchName, targets the first two challenges: it compares data-driven models with seven operational centers under one protocol against GPM IMERG satellite observations~\cite{huffman2020imerg}, introduces the Amplitude-aware Gerrity Skill Score (\AGSS) for anomaly strength and the Rain-belt Latitude Error (\RBLE) for rain-belt location beyond ACC and RMSE, and supplies simulated months from 52 CMIP6 climate models for training~\cite{eyring2016cmip6}.
\looseness=-1

Based on this benchmark, we develop \MODEL, a seasonal precipitation forecasting model that targets the third challenge through sim-to-real transfer: pretraining on these simulated months, then fine-tuning on Earth data (\cref{fig:arch}).
Three properties of the observed record limit this transfer: non-stationarity, sample scarcity, and noisy targets.
\MODEL{} counters each with one module: SimForced corrects the lagging background climate using simulations that run beyond the present, SimMemory learns the effect of earlier seasons from over 25,000 simulated years, and SimAnalog averages the outcomes of similar simulated states to reduce label noise.

Our contributions are summarized as follows:
\begin{itemize}[leftmargin=*, itemsep=1pt, topsep=2pt]
  \item We build \BenchName, a benchmark for global seasonal precipitation forecasting that provides seven operational references, 310,944 simulated training months, and two operational metrics, \AGSS{} and \RBLE{}.
  \item Based on this benchmark, we propose \MODEL, integrating three modules that address non-stationarity, sample scarcity, and noisy targets, respectively, collectively improving its ability to learn from simulations what the observed record lacks.
  \item Extensive experiments show that \MODEL{} outperforms operational systems and data-driven baselines across deterministic, probabilistic, and operational metrics, opening a path for transferring simulation knowledge to real-world Earth system prediction.
\end{itemize}

%% file: sec/2_related.tex
\section{Related Work}
\label{sec:related}\label{sec:related_seasonal}\label{sec:related_bench}\label{sec:related_sim}

\paragraph{Weather forecasting and benchmarks.}
Earthformer~\cite{gao2022earthformer} models space--time interactions for precipitation nowcasting, while AlphaPre~\cite{lin2025alphapre} separates intensity and displacement through amplitude and phase.
Generative approaches capture uncertain local evolution through knowledge-guided latent diffusion~\cite{gao2023prediff}, deterministic--probabilistic cascades~\cite{gong2024cascast}, and residual diffusion~\cite{yu2024diffcast}.
For global weather, SwinRDM~\cite{chen2023swinrdm} combines recurrent forecasts with diffusion super-resolution, OneForecast~\cite{gao2025oneforecast} couples global and regional grids, FengWu-4DVar~\cite{xiao2024fengwu4dvar} integrates data assimilation, and WeatherGFT~\cite{xu2024weathergft} combines physical evolution with neural correction across temporal scales.
WeatherGFM~\cite{zhao2025weathergfm} unifies forecasting and other weather tasks through in-context learning.
\looseness=-1

Operational systems such as ECMWF SEAS5 forecast with coupled Earth-system models~\cite{johnson2019seas5,weber2026seas5bcsd}, while data-driven methods cover regional precipitation~\cite{gibson2021training,pinheiro2025telnet} or leads up to 42 days~\cite{chen2024fuxis2s}.
Closest to ours, Palma et al.~\cite{palma2026datadriven} train on CMIP6 simulations and predict global three-month means, rivaling SEAS5 in some regions; we include their model as a baseline, and go beyond it by forecasting monthly global fields one to five months ahead against seven operational systems on anomaly strength and rain-belt location, and by adapting the simulation-trained model to Earth observations.
Existing benchmarks (\cref{tab:bench}) target medium-range or subseasonal forecasts~\cite{rasp2020weatherbench,rasp2024weatherbench2,dewitt2021rainbench,hwang2019rodeo,mouatadid2023subseasonalclimateusa,vitart2022s2sai,nathaniel2024chaosbench}, regional seasonal forecasts trained on reanalysis~\cite{chen2025seasonbench}, or emulation of emission scenarios~\cite{kaltenborn2023climateset}; they verify against reanalysis or report only deterministic and probabilistic scores, gaps that \BenchName{} closes.

\paragraph{Learning from climate simulations.}
Sim-to-real transfer, training on synthetic data and adapting to real data, is common in vision and robotics~\cite{richter2016playing,tobin2017domain}.
In climate forecasting, prior work trains seasonal predictors on simulations~\cite{gibson2021training}, adapts a climate-model embedding to observations~\cite{pan2022improving}, or pretrains a foundation model on simulations~\cite{nguyen2023climax}, treating simulations as a larger training set for one backbone.
\MODEL{} addresses three properties of the observed record through separate modules, building on model analogs~\cite{ding2018model} and trend correction~\cite{weber2026seas5bcsd} estimated from simulations rather than the short record.
\looseness=-1

%% file: sec/3_benchmark.tex
\section{\BenchName}
\label{sec:bench}

\begin{metricsfig}
  \centering
  \includegraphics[width=\textwidth]{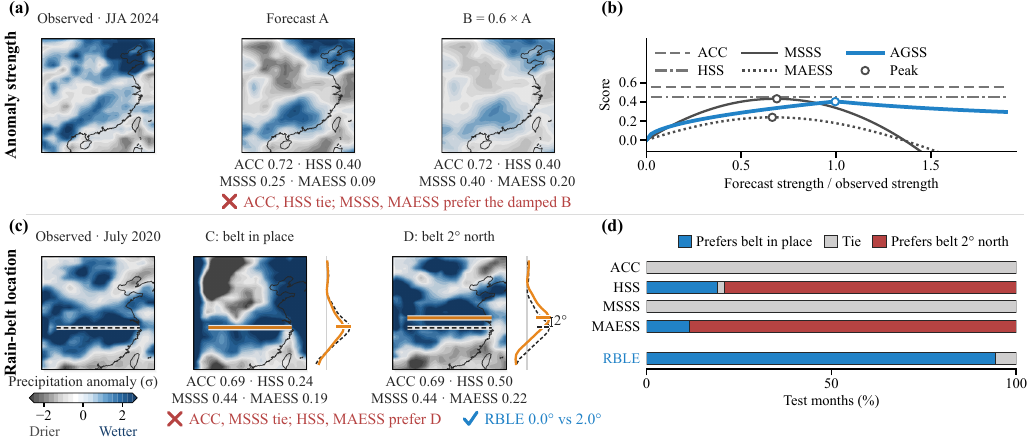}
  \caption{Standard scores miss anomaly strength and rain-belt location. (a)~B damps A by 40\%. (b)~Scores versus forecast strength, 2020--2025 flood seasons. (c)~C and D share ACC and MSSS; D shifts the belt $2^\circ$ north. Strips: zonal-mean anomaly; dashed, observed belt; orange, forecast belt. (d)~Score preferences for matched synthetic pairs in 52 of the 55 2020--2025 test months with an identifiable belt. Map ACC is spatial.}
  \label{fig:metrics}
\end{metricsfig}

\BenchName{} defines a global seasonal forecasting task, gathers the data needed to address it, and evaluates forecasts with standard and operational metrics.

\subsection{Task and Data}
\label{sec:bench_task}\label{sec:bench_data}\label{sec:bench_sim}\label{sec:bench_ops}

We denote by $\mathbf{X}_{<t}$ the oceanic and atmospheric fields observed before an issue month $t$, and by $P_\tau\in\mathbb{R}^{180\times360}$ the monthly precipitation of month $\tau$ on a global $1^\circ$ grid.
A forecaster maps the observed past to the next five months, at leads $\ell=1,\dots,5$:
\begin{equation}
  f:\ \mathbf{X}_{<t}\ \mapsto\ \big(\widehat P_{t+1},\dots,\widehat P_{t+5}\big).
  \label{eq:task}
\end{equation}
Forecasts are verified against GPM IMERG satellite precipitation~\cite{huffman2020imerg} through anomalies $A_\tau=P_\tau-\mu_{m(\tau)}$ from its 1998--2015 climatology $\mu_m$ of calendar month $m(\tau)$.
The test set has 345 issue--lead pairs from January 2020 with targets through December 2025; earlier years may support training and validation, simulations only training.
\looseness=-1

\paragraph{Data.}
GPM IMERG precipitation (1998--2025) is the target because it observes rainfall directly, and the predictors pair the slowly varying ocean, the main source of seasonal predictability, with the recent atmosphere: ERSST v5 SST~\cite{huang2017ersstv5}, GODAS sea-surface height~\cite{behringer2004godas}, and ERA5 precipitation, 2-m temperature and sea-level pressure~\cite{hersbach2020era5}, all on a common $1^\circ$ grid.
Climate simulations add training samples that this single record lacks: we collect 167 runs of 52 CMIP6 models over 1851--2014~\cite{eyring2016cmip6}, of which 158 provide 310,944 training months and nine are held out, extended beyond 2014 by SSP2-4.5 runs of 38 models.
As operational references, we include the monthly ensembles of seven Copernicus centers~\cite{c3s2018seasonal} for 2020--2025, bias-corrected with their hindcasts.
\looseness=-1

\subsection{Evaluation Metrics}
\label{sec:bench_std}\label{sec:bench_metrics}

Standard scores use cell-wise anomalies averaged over leads 1--5; skill scores compare with climatology (\suppref{sec:suppl_metrics}).
\paragraph{Deterministic metrics.} ACC measures forecast--observation correlation across years per cell and calendar month, rewarding sign and pattern independently of amplitude; RMSE and its skill score MSSS measure squared error.
\looseness=-1
\paragraph{Probabilistic metrics.} CRPS generalizes absolute error to a predictive distribution, and RPS scores the probabilities of below-, near- and above-normal terciles used in operational outlooks; CRPSS and RPSS are their skill scores.
\looseness=-1

\begin{maintable}
    \caption{Comparison with statistical, operational and data-driven baselines. Deterministic and probabilistic scores average leads 1--5; operational scores use flood-season totals and monthly rain belts. RMSE and CRPS are in mm/month, RMSE$_{\rm fs}$ in mm and \RBLE{} in degrees. Best (statistical references excluded): \textbf{bold} on {\setlength{\fboxsep}{1pt}\colorbox{bestblue}{blue}}; second: \underline{underlined} on {\setlength{\fboxsep}{1pt}\colorbox{secondblue}{lighter blue}}. $^{\ast}$Pretrained on our simulations and fine-tuned on observations; \textemdash: undefined.}
    \label{tab:main}
    \centering
    \begin{small}
        \begin{sc}
            \renewcommand{\multirowsetup}{\centering}
            \setlength{\tabcolsep}{2.8pt}
            \begin{SRFitTable}{\linewidth}
            \begin{tabular}{l|ccc|ccc|ccccc}
                \toprule
                \multirow{4}{*}{\textbf{Model}} & \multicolumn{11}{c}{\textbf{Metric}} \\
                \cmidrule(lr){2-12}
                & \multicolumn{3}{c|}{\textbf{Deterministic}} & \multicolumn{3}{c|}{\textbf{Probabilistic}} & \multicolumn{5}{c}{\textbf{Operational}} \\
                \cmidrule(lr){2-12}
                & ACC $\uparrow$ & MSSS $\uparrow$ & RMSE $\downarrow$ & CRPS $\downarrow$ & CRPSS $\uparrow$ & RPSS $\uparrow$ & AGSS $\uparrow$ & RMSE$_{\rm fs}$ $\downarrow$ & RBLE $\downarrow$ & PSS $\uparrow$ & RBLSS $\uparrow$ \\
                \midrule
                Climatology  & \textemdash & 0.0000 & 63.98 & 26.58 & 0.0000 & 0.0000 & 0.0000 & 113.0 & 3.41 & 0.000 & 0.000 \\
                Persistence~\cite{vandendool2007empirical}  & 0.0201 & $-$0.8396 & 86.78 & 36.15 & $-$0.3598 & $-$0.3436 & 0.0079 & 133.7 & 4.52 & 0.018 & $-$0.327 \\
                Model analog~\cite{ding2018model}   & 0.1162 & 0.0125 & 63.58 & 26.19 & 0.0146 & $-$0.0166 & 0.0169 & 116.0 & 4.61 & 0.100 & $-$0.353 \\
                \midrule
                ECMWF SEAS5~\cite{johnson2019seas5}   & \cellcolor{secondblue}\underline{0.1358} & \cellcolor{secondblue}\underline{0.0868} & \cellcolor{secondblue}\underline{61.14} & 30.46 & $-$0.1457 & $-$0.2871 & \cellcolor{secondblue}\underline{0.0441} & \cellcolor{secondblue}\underline{111.1} & \cellcolor{secondblue}\underline{3.94} & \cellcolor{secondblue}\underline{0.175} & \cellcolor{secondblue}\underline{$-$0.156} \\
                UKMO GloSea6~\cite{xavier2023assessment}   & 0.1170 & 0.0364 & 62.81 & 33.45 & $-$0.2586 & $-$0.3143 & 0.0291 & 116.0 & 3.96 & 0.092 & $-$0.163 \\
                NCEP CFSv2~\cite{saha2014ncep}   & 0.0977 & 0.0006 & 63.96 & 35.70 & $-$0.3431 & $-$0.2842 & 0.0304 & 118.8 & 4.03 & 0.109 & $-$0.181 \\
                M\'et\'eo-France System 9~\cite{specq2024documentation}   & 0.0968 & $-$0.0145 & 64.44 & 34.58 & $-$0.3010 & $-$0.3215 & 0.0312 & 119.7 & 4.22 & 0.154 & $-$0.238 \\
                DWD GCFS2.2~\cite{frohlich2020gcfs}   & 0.0703 & $-$0.0146 & 64.45 & 35.53 & $-$0.3366 & $-$0.2834 & 0.0190 & 116.3 & 4.31 & 0.052 & $-$0.264 \\
                CMCC SPS4~\cite{sanna2017cmcc}   & 0.1089 & $-$0.0037 & 64.10 & 33.65 & $-$0.2658 & $-$0.3428 & 0.0344 & 126.5 & 4.35 & 0.068 & $-$0.277 \\
                ECCC CanSIPSv3~\cite{diro2024cansips}   & 0.1140 & 0.0230 & 63.24 & 36.81 & $-$0.3848 & $-$0.3688 & 0.0306 & 116.0 & 4.91 & 0.125 & $-$0.439 \\
                \midrule
                ViT$^{\ast}$~\cite{dosovitskiy2021vit}   & 0.1342 & 0.0354 & 62.84 & \cellcolor{secondblue}\underline{25.86} & \cellcolor{secondblue}\underline{0.0271} & \cellcolor{secondblue}\underline{$-$0.0044} & 0.0282 & 117.3 & 4.32 & 0.161 & $-$0.268 \\
                U-Net$^{\ast}$~\cite{ronneberger2015unet}   & 0.1321 & 0.0279 & 63.09 & 25.96 & 0.0235 & $-$0.0084 & 0.0265 & 115.9 & 4.28 & 0.030 & $-$0.256 \\
                GraphCast$^{\ast}$~\cite{lam2023learning}  & 0.0524 & $-$0.0186 & 64.58 & 26.51 & 0.0025 & $-$0.0279 & 0.0006 & 115.8 & 4.08 & $-$0.001 & $-$0.197 \\
                SEASGen~\cite{palma2026datadriven}   & 0.1262 & $-$0.0827 & 66.58 & 36.47 & $-$0.3720 & $-$0.6224 & 0.0064 & 144.6 & 4.82 & 0.011 & $-$0.413 \\
                ClimaX$^{\ast}$~\cite{nguyen2023climax}   & 0.1232 & 0.0074 & 63.75 & 26.16 & 0.0161 & $-$0.0153 & 0.0210 & 116.6 & 4.12 & 0.056 & $-$0.207 \\
                FuXi-S2S~\cite{chen2024fuxis2s}   & 0.1124 & $-$0.1085 & 67.34 & 30.46 & $-$0.1458 & $-$0.2895 & 0.0103 & 169.7 & 4.25 & 0.022 & $-$0.246 \\
                \midrule
                                \MODEL{} (ours)   & \cellcolor{bestblue}\textbf{0.1521} & \cellcolor{bestblue}\textbf{0.0911} & \cellcolor{bestblue}\textbf{61.00} & \cellcolor{bestblue}\textbf{24.92} & \cellcolor{bestblue}\textbf{0.0627} & \cellcolor{bestblue}\textbf{0.0619} & \cellcolor{bestblue}\textbf{0.0454} & \cellcolor{bestblue}\textbf{111.0} & \cellcolor{bestblue}\textbf{3.93} & \cellcolor{bestblue}\textbf{0.188} & \cellcolor{bestblue}\textbf{$-$0.152} \\
                \midrule
                \MODEL{} (Promotion) & 12.00\% & 0.47\% & 0.23\% & 3.63\% & 3.66\% & 6.60\% & 2.95\% & 0.09\% & 0.25\% & 1.58\% & 0.35\% \\
                \bottomrule
            \end{tabular}
            \end{SRFitTable}
        \end{sc}
    \end{small}
\end{maintable}

\subsubsection{What Standard Metrics Miss}
\label{sec:bench_miss}

Beyond the sign and pattern of anomalies, a seasonal outlook states how strongly each region departs from normal and where the monthly rain belt lies~\cite{liu2021precursory}.
To test whether standard scores capture these two properties, we build pairs of forecasts from IMERG observations that differ in only one of them and compare ACC, MSSS, the Heidke (HSS)~\cite{heidke1926berechnung} and MAE (MAESS)~\cite{jolliffe2012forecast} skill scores on each pair (\cref{fig:metrics}).
(i)~\emph{Standard scores can reward damped anomalies.} Damping a forecast leaves ACC and HSS unchanged but raises MSSS and MAESS, whose optima are damped conditional means or medians~\cite{gneiting2011making} (\cref{fig:metrics}a,b).
(ii)~\emph{Standard scores can prefer a displaced rain belt.} Of two forecasts with equal ACC and MSSS, HSS and MAESS mostly prefer the one that moves the belt $2^\circ$ north over the one that keeps it in place (\cref{fig:metrics}c,d), and none states the displacement in degrees.
We therefore add one metric per property, both removing amplitude through the forecast-to-observed amplitude ratio $a$: \AGSS{} scores strength, and \RBLE{} scores location independently of it.

\subsubsection{Anomaly Strength: AGSS}
\label{sec:bench_agss}

The Amplitude-aware Gerrity Skill Score (\AGSS{}) follows principles from the Gerrity score~\cite{gandin1992equitable,gerrity1992note}, an equitable score for ordered categories that rewards correct grades, rare ones more, and gives zero to random or constant forecasts.
It scores flood-season totals over global land, each cell's three wettest consecutive months in the 1998--2015 IMERG climatology, issued $k=1,2,3$ months ahead.
For each $k$, the forecast is graded after its anomalies are divided by the forecast-to-observed amplitude ratio $a_k$, so that the grade score $G_k$ ignores amplitude, and amplitude is then charged separately:
\begin{equation}
  \begin{gathered}
    S_k=\begin{cases}G_k\min\!\left(a_k,a_k^{-1}\right)^{1/2}, & G_k\ge0,\\ G_k, & G_k<0,\end{cases}\\
    \mathrm{AGSS}=\tfrac{1}{3}\textstyle\sum_{k=1}^{3}S_k .
  \end{gathered}
  \label{eq:agss}
\end{equation}
Positive skill is kept in full at the observed amplitude and reduced to 71\% at half or twice it, so \AGSS{} peaks at the observed strength (\cref{fig:metrics}b).
Because inflating a forecast can raise \AGSS{} at an accuracy cost, we report it with $\mathrm{RMSE}_{\mathrm{fs}}$ on the same totals: a system improves when \AGSS{} rises and $\mathrm{RMSE}_{\mathrm{fs}}$ does not.

\subsubsection{Rain-Belt Location: RBLE}
\label{sec:bench_rble}

The Rain-belt Latitude Error (\RBLE{}) follows principles from the mean absolute error, applied to the latitude of the main wet band in each of seven monsoon domains~\cite{yim2014comparison}.
Monthly anomalies are averaged along longitude into a latitude profile (strips in \cref{fig:metrics}c), and the belt lies at $\phi^*$, the squared-anomaly-weighted mean latitude of the positive segment around the profile maximum.
For each case $c$ whose observed month has an identifiable belt, the error is the distance between the forecast and observed belts; a forecast that draws no belt states no location and is charged the error of the 1998--2015 median latitude $\bar\phi_{g,j}$ of domain $g$ and season month $j$:
\begin{equation}
  \begin{gathered}
    e_c=\begin{cases}
      |\phi^*_{f,c}-\phi^*_{o,c}|, & \text{detected},\\
      r_c=|\bar\phi_{g,j}-\phi^*_{o,c}|, & \text{missed},
    \end{cases}\\
    E_g=\frac{\sum_{c\in g}\omega_c e_c}{\sum_{c\in g}\omega_c},\qquad
    R_g=\frac{\sum_{c\in g}\omega_c r_c}{\sum_{c\in g}\omega_c},\\
    \mathrm{RBLE}=\tfrac{1}{7}\textstyle\sum_{g}E_g,\qquad
    \mathrm{RBLSS}=1-\frac{\sum_g E_g}{\sum_g R_g},
  \end{gathered}
  \label{eq:rble}
\end{equation}
where $\omega_c$ are fixed case weights; RBLSS compares \RBLE{} with this climatological position as MSSS does with climatology~\cite{murphy1988skill}.
\RBLE{} prefers the correctly placed belt in almost all pairs of \cref{fig:metrics}(d), and we report beside it the Peirce skill score $\mathrm{PSS}=H-F$~\cite{peirce1884numerical,jolliffe2012forecast}, the belt detection rate minus the false-alarm rate; identifiability, detection and weights are defined in \suppref{sec:suppl_metrics}.

%% file: sec/4_method.tex
\section{\MODEL}
\label{sec:method}

\MODEL{} addresses the third challenge of \cref{sec:intro}, learning from a single, short observational record, through three modules: SimForced, SimMemory, and SimAnalog.
As shown in \cref{fig:arch}, the Earth inputs, the preceding three months of atmospheric and oceanic fields, are expressed as anomalies against a background aligned by SimForced; a U-Net with a Transformer bottleneck maps them to decoder features, to which SimMemory adds an encoding of earlier seasons learned from simulations; and a linear head, a calibration layer and a probability head turn these features into 51-member forecasts for all leads; SimAnalog acts only in training, supplementing the Earth labels (\cref{sec:method_train}).

\subsection{Forecast Decomposition}
\label{sec:method_problem}

For an issue month $t$, \MODEL{} forecasts precipitation at leads $\ell=1,\dots,5$ from the recent state $X_t$, the preceding three months, and the history $\mathcal H_t$ of earlier seasons.
Following the standard decomposition in seasonal forecasting~\cite{palmer1994prospects,doblasreyes2013seasonal}, a variable at cell $c$ and target month $\tau=t+\ell$ is a slowly forced background $\mu$ plus a predictable response $r$ and weather noise $\varepsilon$, scaled by the interannual standard deviation $\sigma$ of calendar month $m(\tau)$:
\begin{equation}
  x_{c,\tau}=\mu_{c,\tau}+\sigma_{c,m(\tau)}\big[r_c(X_t,\mathcal H_t)+\varepsilon_{c,\tau}\big].
  \label{eq:decomp}
\end{equation}
The backbone learns $r$ from the observed $X_t$, but the record poorly determines three parts of it, each because of one property of \cref{sec:intro}: (i)~\textit{non-stationarity}: $\mu$ estimated from past years lags the trend; (ii)~\textit{sample scarcity}: a few decades are too few to learn the effect of $\mathcal H_t$; (iii)~\textit{noisy targets}: each observed label carries its own $\varepsilon$.
Simulations lack these limits, so each of the next three modules estimates one of these parts from simulations and transfers it to Earth.

\subsection{SimForced: Forced Background}
\label{sec:method_simforced}

To estimate the background $\mu$ beyond the end of the observed record, we propose SimForced, which measures in simulations how far a background computed from past years, the only one available on Earth, lags the background centered on the target month, and removes this lag from the Earth anomalies:
\begin{align}
  \Delta_u &= \frac{1}{|\mathcal S_u|}\sum_{s\in\mathcal S_u}\frac{C_{s,u}-T_{s,u}}{\sigma_{s,m(u)}},
  \label{eq:simforced}\\
  \widetilde z^{E}_u &= \frac{x^E_u-T^E_u}{\sigma^E_{m(u)}}-\Delta_u,
  \label{eq:simforced_earth}
\end{align}
where $C_{s,u}$ and $T_{s,u}$ are the 31-year same-calendar-month backgrounds centered on and trailing month $u$ in scenario-extended simulation $s\in\mathcal S_u$, with interannual standard deviation $\sigma_{s,m(u)}$, and superscript $E$ marks the Earth counterparts; the aligned anomaly $\widetilde z^E_u$ follows the centered convention of the simulations and serves as Earth input and target. $\Delta_u$ is a fixed statistic, not a learned quantity; of the three modules, only SimMemory adds trainable parameters.
\looseness=-1

\subsection{SimMemory: Seasonal History}
\label{sec:method_simmemory}

To learn how earlier seasons shape the outcome from more samples than the observed record holds, we propose SimMemory, which pretrains a history encoder on simulations and keeps it frozen on Earth:
\begin{equation}
  \widehat Z_t = H_\psi\big(F_\theta(X_t)+G_\phi(B_t,A_t,\Omega)\big),
  \label{eq:simmemory}
\end{equation}
where $B_t$ holds four six-month means of the aligned fields over lags 4--27, summarizing $\mathcal H_t$; $F_\theta$, $H_\psi$ and $G_\phi$ are the backbone, linear head and memory encoder, $A_t$ marks available blocks, $\Omega$ holds static fields, and $\widehat Z_t$ is the anomaly forecast for all leads.

\subsection{SimAnalog: Analog Outcomes}
\label{sec:method_simanalog}

To reduce the weather noise $\varepsilon$ carried by each single observed outcome, we propose SimAnalog, which retrieves simulated states similar to each Earth training state and blends the average of their outcomes into the Earth label:
\begin{align}
  y_{t\ell c} &= (1-W_{t\ell c})\,\widetilde z^{E}_{t+\ell,c}+W_{t\ell c}\,\underbrace{\tfrac{1}{K}\textstyle\sum_{i=1}^{K} z^{(i)}_{\ell c}}_{M_{t\ell c}},
  \label{eq:simanalog_label}\\
  \mathcal L &= \operatorname{mean}_{t,\ell,c}\; w_c\big[\widehat Z_{t\ell c}-\operatorname{clip}(y_{t\ell c},-5,5)\big]^2,
  \label{eq:simanalog}
\end{align}
where the $K=16$ analogs are same-calendar-month simulated issues whose SST and SSH states are closest to the Earth state at issue $t$, so retrieval needs no Earth precipitation; $z^{(i)}_{\ell c}$ is the anomaly that analog $i$ develops at lead $\ell$ and cell $c$, $W\in[0,0.9]$ grows with the agreement among analogs, and $w_c$ is the cosine-latitude weight.
Averaging over analogs cancels much of their weather noise, so the teacher $M$ approximates the predictable response $r$ of \cref{eq:decomp}.

\subsection{Three-Stage Training}
\label{sec:method_train}

\MODEL{} is trained from simulations to reanalysis to satellite observations, so that the parts estimated from simulations are in place before the observed record is used.
(1)~\textit{Simulation pretraining:} the backbone and head are trained on the 158 training simulations with the loss of \cref{eq:simanalog} on simulated targets, and SimMemory is then trained with both frozen.
(2)~\textit{Earth fine-tuning:} with SimMemory frozen, the backbone and head are fine-tuned on 1979--2015 with the SimAnalog labels of \cref{eq:simanalog_label} on ERA5 precipitation, whose record is about twice as long as that of IMERG.
(3)~\textit{Satellite fine-tuning:} the decoder and head are fine-tuned on IMERG targets of 1998--2012, adding zonal-profile and variance-preservation terms to the loss.
At inference, the forecast anomalies are converted to precipitation with the IMERG climatology, clipped at zero, and passed to a standard additive calibration and probability head fitted on 2016--2019, which issue 51 quantile members.

%% file: sec/5_experiments.tex
\setcounter{dbltopnumber}{1}
\section{Experiments}
\label{sec:exp}

\begin{leadsfig}
  \centering
  \includegraphics[width=\textwidth]{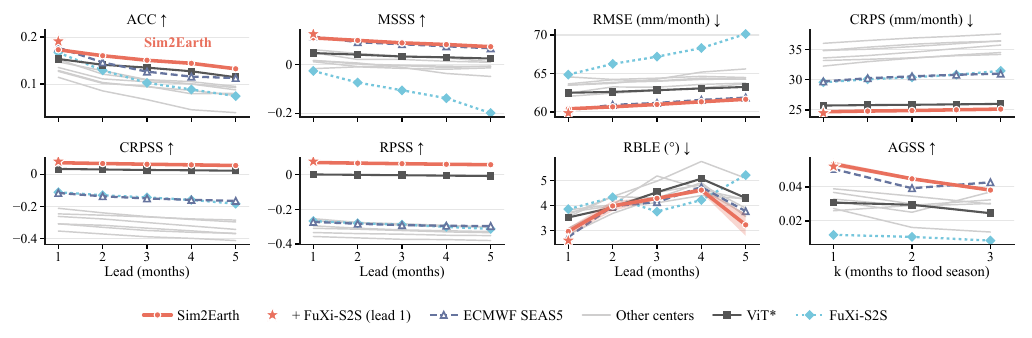}
  \caption{Skill by lead on the 2020--2025 test cases. \MODEL{}: mean of the three seed groups of \cref{tab:main} (shading: $\pm1$ s.d.); star: FuXi-S2S added at lead 1 only. \AGSS{} is plotted against the months $k$ from issue to flood-season start; centers are hindcast-corrected.}
  \label{fig:leads}
\end{leadsfig}

\begin{mechfig}
  \centering
  \includegraphics[width=\textwidth]{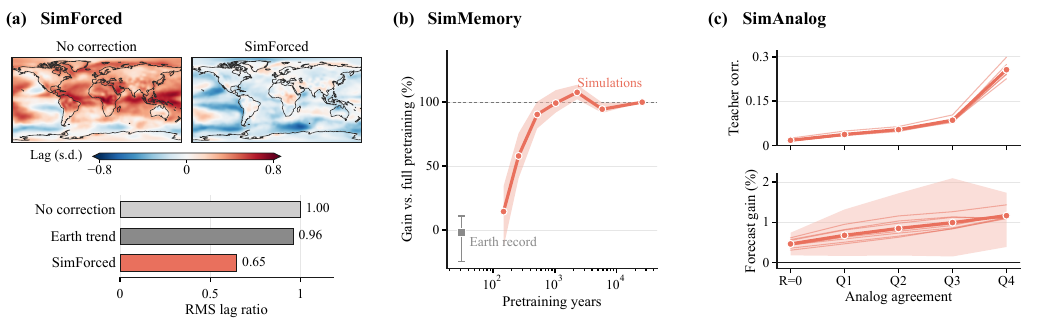}
  \caption{Mechanisms of the three modules. (a)~Background lag (s.d.) of Earth 2-m temperature over 1994--2010 without and with SimForced (smoothed), and its RMS ratio to no correction. (b)~Memory gain versus pretraining years, relative to full pretraining; square: Earth record; band and bar: eight seeds. (c)~Cells binned by analog agreement $R$: teacher correlation with GPM IMERG over 1998--2009 (top; thin: leads) and RMSE reduction from SimAnalog over 2020--2025 (bottom; thin: seeds; band: 95\% interval).}
  \label{fig:mech}
\end{mechfig}

\SetPinHook{7}{\suppressfloats[t]%
\begin{table}[t]
  \caption{Ablation of ACC on the 2020--2025 test period (stage 2; eight seeds, three for ViT and U-Net). $\Delta$ACC is relative to all three modules on the same backbone; ``$\rightarrow$ Earth'': the module is fed an Earth-only estimate.}
  \label{tab:ablation}
  \vspace{-3pt}
  \centering
  \begin{small}
  \begin{sc}
  \setlength{\tabcolsep}{4pt}
  \begin{tabular}{l|cc}
    \toprule
    Variants & ACC $\uparrow$ & $\Delta$ACC \\
    \midrule
    \MODEL{} w/o all modules (Base) & 0.1467 & $-7.0\%$ \\
    Base + SimForced & 0.1499 & $-4.9\%$ \\
    Base + SimMemory & 0.1523 & $-3.4\%$ \\
    Base + SimAnalog & 0.1490 & $-5.5\%$ \\
    \midrule
    \MODEL{} w/o SimForced & 0.1544 & $-2.0\%$ \\
    \MODEL{} w/o SimMemory & 0.1524 & $-3.3\%$ \\
    \MODEL{} w/o SimAnalog & 0.1553 & $-1.5\%$ \\
    \midrule
    SimForced $\rightarrow$ Earth & 0.1546 & $-1.9\%$ \\
    SimMemory $\rightarrow$ Earth & 0.1523 & $-3.4\%$ \\
    SimAnalog $\rightarrow$ Earth & 0.1562 & $-0.9\%$ \\
    \MODEL{} w/o simulations & 0.0540 & $-65.7\%$ \\
    \midrule
    ViT backbone w/o all modules & 0.1401 & $-7.4\%$ \\
    ViT backbone + three modules & 0.1513 & --- \\
    U-Net backbone w/o all modules & 0.1430 & $-8.3\%$ \\
    U-Net backbone + three modules & 0.1559 & --- \\
    \midrule
    \MODEL{} & \textbf{0.1576} & --- \\
    \bottomrule
  \end{tabular}
  \end{sc}
  \end{small}
  \vspace{-2mm}
\end{table}}
\SetPinHook{8}{\suppressfloats[t]%
\begin{figure}[t]
  \centering
  \includegraphics[width=\linewidth]{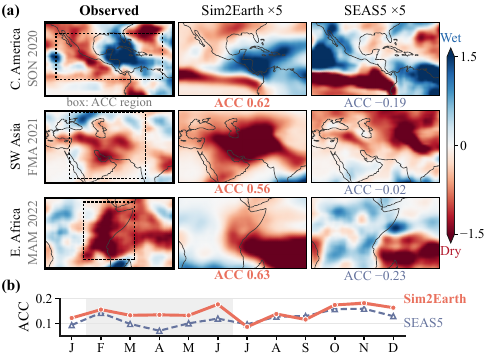}
  \caption{Case study, \MODEL{} seed group A. (a)~Anomalies ($5^\circ$ smoothed) in three La Ni\~na extremes, leads 3--5; ACC over land in the box. (b)~ACC by target month, leads 3--5; grey: February--June.}
  \label{fig:case}
\end{figure}}

\subsection{Experimental Settings}
\label{sec:exp_setup}\label{sec:exp_protocol}

\paragraph{Implementation details.}
\MODEL{} is trained through the three stages of \cref{sec:method_train} with Earth data up to 2015 and calibration on 2016--2019.
One product averages three independently trained seeds, and \cref{tab:main} reports the mean score of three such products with disjoint seeds.

\paragraph{Baselines and evaluation.}
We compare with (1)~statistical references; (2)~seven operational centers, with hindcast-corrected ensemble means and native ensembles; (3)~AI backbones pretrained on our simulations and fine-tuned on observations ($^{\ast}$); and (4)~published ensemble AI models.
The starred backbones use the same 158 training simulations as \MODEL{}; point forecasts without native ensembles use our probability head.
All systems follow the protocol of \cref{sec:bench_task}, and Promotion is the improvement (\suppref{sec:suppl_metrics}) over the strongest operational or data-driven model in each column.

\subsection{Quantitative Results Analysis}
\label{sec:exp_main}

\subsubsection{Comparison with Operational Centers}
\label{sec:exp_ops}

\paragraph{Deterministic skill.}
In \cref{tab:main}, \MODEL{} has the highest ACC, 12.0\% above ECMWF SEAS5, the strongest of the seven centers.
It also has the lowest squared error (RMSE 61.00 vs.\ 61.14\,mm/month for SEAS5), where no other system reaches half of their MSSS.

\paragraph{Probabilistic skill.}
Outlooks are issued as tercile probabilities, and here the native ensembles of all seven centers score below climatology (negative CRPSS and RPSS).
\MODEL{} has the lowest CRPS (24.92 vs.\ 30.46\,mm/month for the native SEAS5 ensemble) and is the only system whose tercile probabilities improve on climatology (RPSS 0.0619).

\paragraph{Operational skill.}
Beyond the pattern, an outlook states anomaly strength and rain-belt location (\cref{sec:bench_miss}).
\MODEL{} attains the highest \AGSS{} (0.0454 vs.\ 0.0441 for SEAS5) at no higher flood-season error, so its low squared error does not come from damping; it also detects the rain belt with the highest skill (PSS 0.188 vs.\ 0.175) at the lowest latitude error (\RBLE{} $3.93^\circ$ vs.\ $3.94^\circ$).
Overall, a data-driven model reaches the level of the leading operational center in all eleven columns (Promotion row).

\subsubsection{Comparison with Data-Driven Models}
\label{sec:exp_dd}\label{sec:exp_simpre}

Existing data-driven models do not reach this level, for three reasons.
(i)~\textit{Simulations without Earth adaptation:} SEASGen, trained on CMIP6, reaches an ACC of 0.1262 but a negative MSSS ($-0.0827$), consistent with the sim-to-real gap that \MODEL{} targets.
(ii)~\textit{Simulations with generic fine-tuning:} the strongest starred model, ViT$^{\ast}$, approaches SEAS5 in ACC (0.1342 vs.\ 0.1358) but not in MSSS (0.0354 vs.\ 0.0868) or \AGSS{} (0.0282 vs.\ 0.0441), a gap that ACC alone would hide.
(iii)~\textit{Models built for shorter ranges:} GraphCast$^{\ast}$ reaches an ACC of only 0.0524, and the autoregressive FuXi-S2S has the largest flood-season error (RMSE$_{\rm fs}$ 169.7\,mm).
Since ViT$^{\ast}$ is pretrained on the same simulations, access to simulations alone does not explain the margin of \MODEL{} (13.3\% in ACC and 61\% in \AGSS{}); \cref{sec:exp_ablation} examines how it is transferred.

\subsubsection{Analysis across Lead Times}
\label{sec:exp_leads}\label{sec:exp_further}

\Cref{fig:leads} resolves the scores over the one to five months of an operational outlook.
The ACC of \MODEL{} decays more slowly than that of SEAS5, and its margin widens to 17--24\% at leads 3--5, consistent with the slow ocean and earlier seasons learned from simulations taking over from the initial state: a five-month outlook from \MODEL{} is more skillful than a three-month one from SEAS5 (ACC 0.133 vs.\ 0.126).

\subsection{Ablation Study}
\label{sec:exp_ablation}\label{sec:exp_abl_modules}\label{sec:exp_mechanisms}\label{sec:exp_samplesize}

\Cref{tab:ablation} tests whether the modules supply what the observed record lacks, retraining every variant with equal budget and seeds.
(1)~\textit{Every module contributes:} removing one lowers ACC by 1.5--3.3\% and all three by 7.0\%, in all eight seeds.
(2)~\textit{The gains add up:} 0.0111 for single additions against 0.0109 jointly, consistent with each module repairing a separate part of \cref{eq:decomp}.
(3)~\textit{The gains come from simulations:} feeding a module an Earth-only estimate keeps its architecture but loses 0.9--3.4\%, so the gain lies mainly in simulated information, not capacity; without simulations ACC falls by 65.7\%.
(4)~\textit{The gains transfer across backbones:} the three modules raise ACC by 0.0112 on a ViT and by 0.0129 on a convolutional U-Net, in all three seeds; \cref{fig:mech} shows how each module works.

\paragraph{SimForced counters non-stationarity.}
Over 1994--2010, the 2-m temperature background computed from past years lies below the centered one over 88\% of the globe (\cref{fig:mech}a).
Subtracting the lag estimated from simulations reduces it over 72\% of the globe and by 35\% in root-mean-square, whereas extrapolating the Earth's own trend removes only 4\% and costs as much ACC as removing SimForced (\cref{tab:ablation}).
Simulations, running beyond the present, thus supply the background the record has not reached.

\paragraph{SimMemory counters sample scarcity.}
Pretrained on the 31 Earth years of 1979--2009, the history encoder yields no gain; pretrained on simulations, its gain appears near 150 years ($5\times$ the record) and saturates near 1,000 ($32\times$ the record, \cref{fig:mech}b).
Seasonal memory thus needs 5--32 times the record, which simulations provide.

\paragraph{SimAnalog counters noisy targets.}
If averaging analogs cancels weather noise, their mean should carry more signal where they agree.
From cells without agreement to the top agreement quartile, teacher--observation correlation rises from 0.02 to 0.26, and SimAnalog's error reduction from 0.46\% to 1.16\% (\cref{fig:mech}c).
This is why $W$ grows with agreement, adding simulated outcomes where they carry signal.
\looseness=-1

\subsection{Case Study}
\label{sec:exp_case}

\paragraph{La Ni\~na extremes, 2020--2023.}
\Cref{fig:case}(a) revisits three precipitation extremes, each forecast three to five months ahead~\cite{wmo2021sogc2020,wmo2022sogc2021,wmo2023sogc2022}.
\MODEL{} reproduces all three patterns (ACC 0.56--0.63), unlike SEAS5 ($-0.23$ to $-0.02$), and has the higher pattern correlation in 61\% of all 459 regional La Ni\~na cases.
In these La Ni\~na extremes, a model trained mainly on simulations anticipates impacts that the leading dynamical system misses.
\looseness=-1

\paragraph{Spring predictability barrier.}
\MODEL{} also keeps skill across the spring predictability barrier~\cite{webster1992monsoon,torrence1998annual} (ACC 0.146 vs.\ 0.106 for SEAS5 at leads 3--5, \cref{fig:case}b), a margin that also holds on the 2016--2019 validation years (0.137 vs.\ 0.098) and with any single test year removed.
\looseness=-1

%% file: sec/6_conclusion.tex
\section{Conclusion}
\label{sec:conclusion}

We present \BenchName{}, the first global benchmark comparing data-driven forecasts with operational centers on anomaly strength and rain-belt location, and \MODEL{}, which learns from simulations what the short record lacks and reaches the level of the leading center on all three score types, each module gaining from the simulated information it transfers---a path from simulated worlds to Earth system prediction.
\looseness=-1

%% file: sec/X_suppl.tex
\clearpage
\setcounter{page}{1}
\renewcommand{\thepage}{S\arabic{page}}
\maketitlesupplementary
\appendix

\input{sec/X_bench}
\clearpage
\input{sec/X_experiments}

%% file: sec/X_bench.tex
\section{\BenchName{} Details}
\label{sec:suppl_bench}

\BenchName{} pairs a global satellite target with three groups of data: observational and reanalysis predictors, climate simulations for training, and operational forecasts for reference.
This section describes each group, the predictor sets built from them, and the processing that places all of them on one grid.

\subsection{Overview and Predictor Sets}
\label{sec:suppl_bench_overview}

Seasonal predictability comes mainly from slowly varying boundary conditions, the ocean, sea ice and land surface, and from how the atmosphere responds to them.
Following SeasonBench-EA~\cite{chen2025seasonbench}, we therefore select variables by two criteria: they characterize the large-scale circulation or a boundary condition, and they are established sources of seasonal precipitation skill.
We add a third criterion that is specific to learning from simulations: a predictor is \emph{paired} when a variable of the same definition is available both on Earth and in the CMIP6 simulations, so that a model can be pretrained on simulations and applied to Earth with the same inputs.
\Cref{tab:bench_vars} lists the three resulting sets:
\begin{itemize}[leftmargin=*, itemsep=1pt, topsep=2pt]
  \item \textbf{Core predictors} (5 variables): precipitation, 2-m temperature, sea-level pressure, sea surface temperature and sea-surface height, the inputs of \MODEL{} and of the simulation-pretrained baselines in \cref{tab:main}.
  \item \textbf{Extended paired predictors} (21 variables): surface state, radiation, surface fluxes, upper ocean and sea ice, each with an observational counterpart and a CMIP6 counterpart.
  \item \textbf{Observation-only predictors}: pressure-level circulation and land-surface fields from reanalysis, for models trained on the observed record alone, as in SeasonBench-EA.
\end{itemize}
All predictors are monthly means on the global $1^\circ$ grid of the target, and the observational fields cover 1979--2025.

\begin{table*}[t]
  \caption{Variables in \BenchName{}. Each paired predictor lists its observational or reanalysis product and its CMIP6 counterpart with the number of the 167 historical runs that provide it; all simulated fields span 1851--2014 and are extended to 2025 with SSP2-4.5 where available. Observation-only predictors have no simulated counterpart in the benchmark.}
  \label{tab:bench_vars}
  \centering
  \begin{SRFitTable}{\textwidth}
\SRTableStyle[3.0pt]
\begin{tabular}{ll|ll|lc}
    \toprule
    Group & Variable & Product & Period & CMIP6 & Runs \\
    \midrule
    Target & Precipitation & GPM IMERG~\cite{huffman2020imerg} & 1998--2025 & pr & 167 \\
    \midrule
    \multirow{5}{*}{Core} & Precipitation & ERA5~\cite{hersbach2020era5} & 1979--2025 & pr & 167 \\
     & 2-m temperature & ERA5 & 1979--2025 & tas & 167 \\
     & Sea-level pressure & ERA5 & 1979--2025 & psl & 167 \\
     & Sea surface temperature & ERSST v5~\cite{huang2017ersstv5} & 1979--2025 & ts (ocean) & 167 \\
     & Sea-surface height & GODAS~\cite{behringer2004godas} & 1980--2025 & zos & 167 \\
    \midrule
    \multirow{6}{*}{Surface state} & Surface pressure & ERA5 & 1979--2025 & ps & 162 \\
     & 10-m wind, eastward / northward & ERA5 & 1979--2025 & uas / vas & 134 \\
     & 10-m wind speed & ERA5 & 1979--2025 & sfcWind & 152 \\
     & Precipitable water & ERA5 & 1979--2025 & prw & 156 \\
     & Total cloud cover & ERA5 & 1979--2025 & clt & 159 \\
     & Surface stress, eastward / northward & ERA5 & 1979--2025 & tauu / tauv & 159 / 158 \\
    \midrule
    \multirow{4}{*}{Radiation} & Outgoing longwave, top of atmosphere & ERA5 & 1979--2025 & rlut & 164 \\
     & Reflected shortwave, top of atmosphere & ERA5 & 1979--2025 & rsut & 158 \\
     & Downward shortwave, surface & ERA5 & 1979--2025 & rsds & 159 \\
     & Downward longwave, surface & ERA5 & 1979--2025 & rlds & 149 \\
    \midrule
    \multirow{3}{*}{Surface fluxes} & Sensible heat flux & ERA5 & 1979--2025 & hfss & 157 \\
     & Latent heat flux & ERA5 & 1979--2025 & hfls & 159 \\
     & Evaporation & ERA5 & 1979--2025 & evspsbl & 161 \\
    \midrule
    \multirow{6}{*}{Ocean and sea ice} & Sea surface temperature (ocean grid) & ERSST v5 & 1979--2025 & tos & 167 \\
     & Sea surface salinity & ORAS5~\cite{zuo2019oras5} & 1979--2025 & sos & 152 \\
     & Mixed-layer depth & ORAS5 & 1979--2025 & mlotst & 76 \\
     & Net surface heat flux into the ocean & ORAS5 & 1979--2025 & hfds & 138 \\
     & Sea-ice concentration & NSIDC CDR v6~\cite{meier2024seaice} & 1979--2025 & siconc & 156 \\
     & Sea-ice thickness & ORAS5 & 1979--2025 & sithick & 128 \\
    \midrule
    \multirow{4}{*}{Observation-only} & $z$, $t$, $u$, $v$, $q$ at 1000, 850, 700, 500, 200\,hPa; $w$ at 850, 500, 200\,hPa & ERA5 & 1979--2025 & \textemdash & \textemdash \\
     & Boundary-layer height, soil temperature, snow albedo, snow water equivalent & ERA5 & 1979--2025 & \textemdash & \textemdash \\
     & Soil moisture (four layers), snow depth, snow cover & ERA5-Land~\cite{munozsabater2021era5land} & 1979--2025 & \textemdash & \textemdash \\
     & Upper-300\,m heat content, 20\,$^\circ$C isotherm depth & ORAS5 & 1979--2025 & \textemdash & \textemdash \\
    \midrule
    Static & Land--sea mask, soil type, surface geopotential & ERA5 & \textemdash & \textemdash & \textemdash \\
    \bottomrule
  \end{tabular}
\end{SRFitTable}
\end{table*}

\subsection{Observation and Reanalysis Data}
\label{sec:suppl_data}

\paragraph{Target.}
GPM IMERG~\cite{huffman2020imerg} is the target because it observes rainfall directly and covers the globe since 1998.
We conservatively remap its $0.25^\circ$ daily fields to $1^\circ$ by spherical area overlap and aggregate calendar-month totals; anomalies are taken from its 1998--2015 calendar-month climatology (\cref{sec:bench_task}).
\looseness=-1

\paragraph{Predictors.}
Atmospheric and land predictors come from ERA5~\cite{hersbach2020era5} and ERA5-Land~\cite{munozsabater2021era5land}, sea surface temperature from ERSST v5~\cite{huang2017ersstv5}, sea-surface height from GODAS~\cite{behringer2004godas}, upper-ocean fields from ORAS5~\cite{zuo2019oras5}, and sea-ice concentration from the NOAA/NSIDC passive-microwave climate data record~\cite{meier2024seaice}.
ERA5 single-level and pressure-level monthly means are retrieved on integer-degree nodes and averaged to the $1^\circ$ cells with area weights; ERA5-Land, ORAS5 and the sea-ice record are remapped by first-order conservative remapping, using sub-cell sampling for the curvilinear ocean and polar-stereographic grids.

\subsection{Climate Simulations}
\label{sec:suppl_bench_sim}

The simulations are the 167 historical runs of 52 CMIP6 models~\cite{eyring2016cmip6} of \cref{sec:bench_task}, 158 for training and nine held out, over 1851--2014, extended to 2025 by the SSP2-4.5 runs of the same members~\cite{oneill2016scenariomip} where available.
For every variable of \cref{tab:bench_vars}, we take the monthly output of each run that provides it, so the number of runs differs by variable; atmospheric fields are remapped conservatively from regular model grids, and ocean and sea-ice fields from curvilinear and unstructured grids with the same sub-cell scheme as above.
Fluxes and stresses are converted to the units and sign conventions of CMIP6 on both sides (upward positive for outgoing radiation and turbulent fluxes, downward positive for the net heat flux into the ocean), so that a paired predictor carries the same quantity in simulations and on Earth.
As for the core predictors, each simulated variable is expressed as anomalies about a centered 31-year background and scaled by its model's interannual standard deviation, which removes the climatological biases of individual models.

\subsection{Operational Centers}
\label{sec:suppl_centers}

\begin{table}[t]
  \caption{The seven operational seasonal forecast systems in \BenchName. Horizontal resolutions (km) are approximated by their mid-latitude values for atmosphere (A), land (L) and ocean (O).}
  \label{tab:centers}
  \centering
  \begin{SRFitTable}{\linewidth}
\SRTableStyle[2.8pt]
\begin{tabular}{l|lcc}
    \toprule
    Center & System & Resolution & Members \\
    \midrule
    ECMWF~\cite{johnson2019seas5} & SEAS5 & A36-L36-O20 & 51 \\
    UK Met Office~\cite{xavier2023assessment} & GloSea6-GC5.1 & A60-L60-O20 & 56 \\
    NCEP~\cite{saha2014ncep} & CFSv2 & A110-L110-O40 & 116 \\
    M\'et\'eo-France~\cite{specq2024documentation} & System 9 & A50-L50-O20 & 51 \\
    DWD~\cite{frohlich2020gcfs} & GCFS2.2 & A100-L100-O40 & 50 \\
    CMCC~\cite{sanna2017cmcc} & CMCC-SPS4 & A50-L50-O28 & 50 \\
    ECCC~\cite{diro2024cansips} & CanSIPSv3.0 & A110-L110-O100 & 20 \\
    \bottomrule
  \end{tabular}
\end{SRFitTable}
\end{table}

Each center issues an ensemble of coupled-model runs from perturbed initial states, and the seven systems of \cref{tab:centers} span ensemble sizes from 20 to 116 and horizontal resolutions from 20 to 110\,km.
For each C3S start month~\cite{c3s2018seasonal}, we use the latest available system version and its hindcasts, and all 72 issue months of 2020--2025 are available for every center.
As in operational practice, the hindcasts remove each center's systematic bias: after conversion to monthly totals, a forecast is corrected by subtracting the center's 1993--2016 hindcast climatology for that start month and lead, adding the observed target-month climatology, and clipping negative totals to zero.
The probabilistic scores use the native members of each center, so that every system is scored as it is issued.

\subsection{Splits and Usage}
\label{sec:suppl_bench_usage}

The test set contains the 345 issue--lead pairs issued from January 2020, with targets through December 2025, at leads 1--5.
Earlier years may be used for training and validation, and simulations only for training; in our experiments, Earth fine-tuning uses 1979--2015, and stage-3 and calibration settings are selected on 2013--2015 and 2016--2019.
The same protocol, grid, climatology and scorer apply to every system in \cref{tab:main}, and the predictor sets of \cref{tab:bench_vars} let methods be compared at matched inputs: with the core predictors, as in \cref{tab:main}, with all paired predictors when pretraining on simulations, or with observation-only predictors when training on the observed record alone.

\subsection{Stability across Seed Groups}
\label{sec:suppl_bench_stability}

\Cref{tab:main} reports the mean of three disjoint three-seed groups of \MODEL{}.
\Cref{fig:leads} shows the spread of the three groups by lead.
\looseness=-1

%% file: sec/X_experiments.tex
\section{Evaluation Metrics}
\label{sec:suppl_metrics}

This section completes the metric definitions of \cref{sec:bench_std}: \cref{sec:suppl_standard_scores} details the standard scores, \cref{sec:suppl_metric_defs} the samples and estimators of \AGSS{} and \RBLE{}, and \cref{sec:suppl_metric_props} the properties on which the two metrics rest, with proofs. Writing $s$ for \MODEL{} and $b$ for the strongest operational or data-driven model in a column, Promotion in \cref{tab:main} is $(s-b)/|b|$ for ACC and \AGSS{}, $(b-s)/b$ for RMSE, CRPS, RMSE$_{\rm fs}$ and \RBLE{}, and $(s-b)/(1-b)$ for the bounded skill scores (MSSS, CRPSS, RPSS, PSS, RBLSS), all computed from the scores as displayed in the table; the skill-score form measures the fraction of remaining headroom to the ceiling of 1 rather than a relative change in the score itself, which is why its percentages are not directly comparable across the two groups of columns.
\AGSS{} is scored on global land cells and \RBLE{} in seven monsoon domains, both against IMERG; target months after December 2025 are excluded, so no 2026 observation is used.

\subsection{Standard Scores}
\label{sec:suppl_standard_scores}

We evaluate both the predicted monthly amount and the uncertainty represented by an ensemble.
Spatial averages of the following gridded scores use latitude-cosine weights.

\noindent\textbf{Deterministic metrics.}
ACC measures how well forecasts reproduce year-to-year precipitation variations at each location.
For each grid cell and target calendar month, we correlate forecasts and observations across years, then average over cells and calendar months.
MSSS expresses squared-error reduction relative to monthly climatology,
$\mathrm{MSSS}=1-\mathrm{MSE}/\mathrm{MSE}_{\mathrm{clim}}$~\cite{murphy1988skill}: zero matches climatology and positive values improve upon it.
RMSE measures the error in monthly totals in mm/month.
We report ACC, MSSS and RMSE at every lead and averaged over leads 1--5; higher ACC/MSSS and lower RMSE are better.

\smallskip\noindent\textbf{Probabilistic metrics.}
CRPS evaluates the full forecast distribution against observed precipitation~\cite{hersbach2000crps}.
RPS evaluates cumulative probabilities for below-normal, near-normal and above-normal rainfall, defined by climatological terciles~\cite{weigel2007rpss}.
Both scores are lower-is-better.
Their skill forms, CRPSS and RPSS, express improvement over a climatological ensemble drawn from 1998--2015, with positive values indicating better performance.

\subsection{Definitions and Estimators}
\label{sec:suppl_metric_defs}

\noindent\textbf{AGSS sample.}
A cell enters the \AGSS{} sample if its land fraction exceeds 0.5 and, in 1998--2015, it has at least 15 finite reference years, a mean flood-season total of at least 10\,mm, a positive lowest category boundary and strictly increasing boundaries.
The local flood season comprises the three consecutive months with the largest 1998--2015 IMERG climatological total.
Reference seasonal totals determine each cell's mean $n_i$, standard deviation $\mathrm{sd}_i$ and five category boundaries.
For issue offset $k$, the forecast is issued $k$ months before the first flood-season month, so the season covers leads $k$ to $k+2$.
The validation years are issue years 2016--2019 and the test years 2020--2025.
A system is scored in a period if at least 90\% of its samples are finite, and all scored systems share the same finite support.

\smallskip\noindent\textbf{AGSS category score.}
For cell $i$, let $n_i$ be the number of valid evaluation years, $J_i$ its observed--forecast joint category frequencies, and $p_i,q_i$ the observed and forecast marginal frequencies.
The Gerrity matrix $S$ is constructed from the pooled observed frequencies.
Define the observed payoff $T=\sum_i n_i\langle J_i,S\rangle$, the forecast chance payoff $B=\sum_i n_i p_i^\top S q_i$, the perfect payoff $P=\sum_i n_i\sum_k p_{ik}S_{kk}$, and its chance payoff $H=\sum_i n_i p_i^\top S p_i$.
The category skill is
\begin{equation}
  G=\frac{T-B}{P-H}.
  \label{eq:gerrity_stratified}
\end{equation}
With cell weights $w_i=\cos\phi_i$, each sum over $i$ is weighted by $w_i$, and the pooled frequencies are weighted in the same way.
$G$ is undefined when the scoring matrix is undefined or $P-H\leq0$.
The forecast is graded after its anomaly has been divided by the amplitude ratio $a$ of \cref{eq:app_amp}, and $G:=0$ when $a=0$.
The amplitude ratio follows the variability component of the Kling--Gupta efficiency~\cite{gupta2009decomposition}.

\smallskip\noindent\textbf{RBLE domains and cases.}
The seven domains use the following longitude and latitude bounds (degrees east and north, with negative values indicating west and south): East Asia $(105,122;22,45)$, South Asia $(70,105;10,30)$, West Africa $(-30,30;5,15)$, North America $(-110,-80;7.5,22.5)$, South America $(-70,-40;-25,-5)$, Australia $(110,150;-20,-5)$, and southern Africa $(25,70;-25,-7.5)$.
The seven domains are adapted from the eight regional monsoon domains of \citet{yim2014comparison}: we omit the western North Pacific domain and set the East Asian domain to $105$--$122^\circ$E and $22$--$45^\circ$N.
East Asia uses June--August; South Asia, West Africa and North America use June--September; the remaining domains use December--March.
Monthly anomalies about the 1998--2015 same-month IMERG normal are converted to mm/day and averaged within $1^\circ$ latitude bands, requiring at least five valid cells per band, and then smoothed with a three-band moving average.
A belt is identifiable under the rule of \cref{sec:suppl_metric_props}; the domain scales $s_g$ are 1.61 (East Asia), 1.56 (South Asia), 0.79 (West Africa), 1.86 (North America), 1.19 (South America), 2.27 (Australia) and 1.46\,mm/day (southern Africa).
A case is a target year, a season month and a lead in 1--5; cases without an identifiable observed belt are excluded from \RBLE{} for every system and enter the false-alarm rate, and each lead receives equal weight.
All systems are scored on the same cases and cells.
On 2016--2019, the NCEP and UKMO hindcasts are incomplete, so these two centers are not ranked in the validation years.

\smallskip\noindent\textbf{RBLE detection, weights and reference.}
A month has an identifiable observed belt when the profile maximum exceeds $s_g$, the root mean square of the domain's 1998--2015 observed profiles; a forecast detects a belt when its maximum exceeds $a_{g,\ell}s_g$, where $a_{g,\ell}$ is its amplitude ratio at lead $\ell$, so a common rescaling changes neither its detections nor its latitudes.
The weights $\omega_c$ cover cases with an identifiable observed belt and give each lead equal weight.
When an observed month has a second identifiable segment whose peak is at least 0.8 of the first (3 of 385 months), the nearer latitude is used.
Scoring a miss at the climatological latitude follows CPC, which scores equal-chances outlooks as zero-skill climatology~\cite{peng2012analysis,peng2013comparison}; such a forecast scores exactly zero, like the reference itself~\cite{murphy1988skill}.
Errors are pooled over domains before the ratio is taken, as in the overall MSSS of the WMO long-range standard~\cite{wmo2007svslrf}, so that domains with a small $R_g$ cannot dominate; the reference has no interannual information and $H=F=1$.
All scores use the issued forecasts without recalibration.

\subsection{Metric Checks}
\label{sec:suppl_metric_checks_main}

Two checks on observations confirm that the operational metrics respond to the property they are designed to score.

\paragraph{Amplitude.}
For noisy synthetic forecasts constructed from IMERG observations, \AGSS{} peaks at amplitude ratios $a=0.99$--$1.01$ for every correlation $\rho$ from 0.06 to 0.96, while MSSS peaks near $a=\rho$.
A fully synthetic version of this check uses a panel of 24 evaluation years, 64 land cells and three issue offsets (random seed 20261103).
An 18-year reference at fixed normal quantiles defines seasonal means of 300\,mm, standard deviations of 40\,mm, and six categories separated by the 10th, 30th, 50th, 70th and 90th percentiles.
For centered, unit-variance observed anomalies $O'$ and orthogonal unit-variance noise $E$, the standardized forecast is $a(0.6O'+0.8E)$, with $a\in[0.2,1.8]$; opposite-sign longitude pairs give zero spatial means.
Each season is split into three equal monthly totals.
We compute temporal ACC and monthly RMSE/MSSS over all cases, and apply the full \AGSS{} estimator of \cref{eq:agss}, including normalized category scoring and the mean over three issue offsets.
All four metrics are evaluated at steps of 0.02 in $a$.

\paragraph{Month order.}
Exchanging the monthly anomaly totals of the first and last season months leaves the seasonal totals, and hence their ACC and RMSE, unchanged, but gives an \RBLE{} of 4.4--9.5$^\circ$ across the seven domains.
Calendar-month ACC can detect this exchange; the invariance applies to seasonal-total scores.

\input{sec/X_metric_props}

%% file: sec/X_metric_props.tex
\subsection{Properties of AGSS and RBLE}
\label{sec:suppl_metric_props}

This subsection states the properties behind the two operational metrics and proves them where possible.
Statements marked \emph{(numerical)} are not proved; we check them on synthetic forecasts, built either in a Gaussian model or from IMERG observations, and no forecast system enters these checks.
Both metrics are global: \AGSS{} is computed over global land cells and \RBLE{} over the seven monsoon domains.
\Cref{tab:app_status} lists every result with its status and assumptions.

\smallskip\noindent\textbf{Gaussian model.}
At one location, let $O'=O-n$ and $F'=F-n$ be anomalies about the reference normal $n$, scaled so that $\operatorname{Var}O'=1$.
Let $Z\sim\mathcal N(0,1)$ be the information available to the forecast, and $O'=\rho Z+\sqrt{1-\rho^2}\,\varepsilon$ with $\varepsilon\sim\mathcal N(0,1)$ independent of $Z$ and $0<\rho<1$.
The forecasts $F'=aZ$ with $a>0$ all have correlation $\rho$ with $O'$ and amplitude ratio $a=\sigma_f/\sigma_o$, so they differ only in amplitude.
We write $\varphi,\Phi$ for the standard normal density and distribution function and $\Phi_2(\cdot,\cdot;\rho)$ for the bivariate one.

\smallskip\noindent\textbf{AGSS as evaluated.}
The sample consists of eligible global land cells $i$ (land fraction above 0.5) with weights $w_i=\cos\phi_i$, their flood-season totals, and three units $u$, one for each issue offset $k=1,2,3$; the target months therefore lie at leads 1--5.
In a unit, a total $X_{it}$ is standardized as $z_{it}=(X_{it}-n_i)/\mathrm{sd}_i$ with the reference mean $n_i$ and standard deviation $\mathrm{sd}_i$.
Subtracting each cell's mean over its valid years and then each year's weighted spatial mean gives $\tilde z_{it}$, and
\begin{equation}
  a_u=\Big(\textstyle\sum_{i,t}w_i\,(\tilde z^{F}_{it})^2\big/\sum_{i,t}w_i\,(\tilde z^{O}_{it})^2\Big)^{1/2}.
  \label{eq:app_amp}
\end{equation}
The forecast is graded after normalization, $\tilde F_{it}=n_i+(F_{it}-n_i)/a_u$, into the six observed categories, and $G_u$ is the stratified Gerrity score of \cref{eq:gerrity_stratified} with cell weights $w_i$; $G_u:=0$ when $a_u=0$.
With an amplitude factor $f$,
\begin{equation}
  S_u=\begin{cases}G_u\,f(a_u), & G_u\ge0,\\ G_u, & G_u<0,\end{cases}
  \qquad \mathrm{AGSS}=\tfrac13\textstyle\sum_uS_u .
  \label{eq:app_agss}
\end{equation}
The reported score uses $f(a)=\min(a,1/a)^{1/2}$.
The results hold for any $f$ with (F1) $f(1)=1$, (F2) $f(a)=f(1/a)$, (F3) $f$ strictly increasing on $[0,1]$, and (F4) $f(0)=0$.
The forms $\min(a,1/a)$, $2a/(1+a^2)$ and $4/(a+1/a)^2$, and their square roots, satisfy (F1)--(F4).

\smallskip\noindent\textbf{RBLE as evaluated.}
In domain $g$, each case $c$ is a target year, a season month and a lead in 1--5.
$P_c(\phi_b)$ is the smoothed zonal-mean monthly anomaly in latitude band $b$ ($1^\circ$ bands, $\phi_1<\dots<\phi_{n_b}$).
Let $P^+=\max(P,0)$, let $m(P)=\max_bP^+(\phi_b)$ be the largest positive value, and let $s_g$ be the root mean square of all finite observed profile values of domain $g$ over the season months of 1998--2015.
The belt is the connected run $\mathcal R(P)$ of positive bands that contains $m(P)$, and its latitude is
\begin{equation}
  \phi^*(P)=\frac{\sum_{b\in\mathcal R(P)}\phi_b\,P(\phi_b)^2}{\sum_{b\in\mathcal R(P)}P(\phi_b)^2}.
  \label{eq:app_belt}
\end{equation}
The location component of SAL also uses a centre of mass~\cite{wernli2008sal}; here it is taken over squared positive anomalies of a latitude profile.
An observed profile has an \emph{identifiable} belt when at least five bands are finite and $m(P^o_c)\ge\theta s_g$, with $\theta=1$.
For a forecast, let $a_{g,\ell}=\big(\sum(P^f)^2/\sum(P^o)^2\big)^{1/2}$, with sums over the cases of domain $g$ at lead $\ell$ whose observed profile has five finite bands and over the bands finite in both profiles.
The forecast \emph{detects} case $c$ when at least five bands are finite, $a_{g,\ell}>0$ and $m(P^f_c)\ge\theta\,a_{g,\ell}\,s_g$; otherwise it misses.
An observed month is \emph{ambiguous} when its second-strongest run is also identifiable and its peak is at least $\kappa=0.8$ times $m(P^o_c)$; let $\phi^{**}_c$ be that run's latitude.
Write $\delta_c(x)=|x-\phi^*(P^o_c)|$, or $\min\big(|x-\phi^*(P^o_c)|,|x-\phi^{**}_c|\big)$ for an ambiguous month.
The case error is $e_c=\delta_c\big(\phi^*(P^f_c)\big)$ for a detection and $e_c=r_c$, the reference error defined below, for a miss.
For season month $j$, let $\bar\phi_{g,j}$ be the median of $\phi^*$ over the reference seasons $y=1998,\dots,2015$ whose observed belt is identifiable.
The reference error is $r_c=\delta_c(\bar\phi_{g,j})$, the error of a fixed belt at that latitude.
A case enters only when its observed belt is identifiable, and each lead receives equal weight; the case weights $\omega_c$ therefore do not depend on the forecast.
Cases whose observed profile has five finite bands but no identifiable belt instead enter the false-alarm rate (FAR), the weighted fraction in which the forecast detects a belt.
Then
\begin{equation}
  \begin{gathered}
  \mathrm{RBLE}_g=\frac{\sum_c\omega_ce_c}{\sum_c\omega_c},\qquad
  \mathrm{RBLSS}_g=1-\frac{\sum_c\omega_ce_c}{\sum_c\omega_cr_c},\\
  \mathrm{RBLSS}=1-\frac{\sum_g E_g}{\sum_g R_g},
  \end{gathered}
  \label{eq:app_rblss}
\end{equation}
where RBLSS, the $R_g$-weighted mean of $\mathrm{RBLSS}_g$, matches the pooled definition of \cref{eq:rble} reported in \cref{tab:main}; $\mathrm{RBLSS}_g$ is used only as an intermediate quantity in the properties below.
Writing $H_g$ for the weighted detection rate and $\mathrm{MAE}_{\mathrm{det},g}$ for the weighted mean of $e_c$ over detections, $\mathrm{RBLE}_g=H_g\,\mathrm{MAE}_{\mathrm{det},g}+(1-H_g)\,\bar r_{\mathrm{miss},g}$, where $\bar r_{\mathrm{miss},g}$ is the weighted mean of $r_c$ over misses and each term is zero when its weight is.

\smallskip\noindent\textbf{Category score.}
\begin{lemma}[Threshold form of the grade score]\label{lem:app_threshold}
For each boundary $\ell=1,\dots,K-1$, let $X^\ell_{it}=\mathbb 1[O_{it}\text{ falls in a category}\le\ell]$, let $Y^\ell_{it}$ be the same indicator for the graded forecast, and let $p_\ell$ be the weighted pooled observed frequency of $X^\ell=1$, from which the Gerrity matrix is built.
With $\widehat{\mathrm{Cov}}_i$ and $\widehat{\mathrm{Var}}_i$ taken over the $n_i$ valid years of cell $i$,
\begin{equation}
  G=\frac{\sum_\ell h_\ell\sum_iw_in_i\,\widehat{\mathrm{Cov}}_i(X^\ell,Y^\ell)}{\sum_\ell h_\ell\sum_iw_in_i\,\widehat{\mathrm{Var}}_i(X^\ell)},\qquad h_\ell=\frac{1}{p_\ell(1-p_\ell)},
  \label{eq:app_threshold}
\end{equation}
which for $w_i=1$ equals \cref{eq:gerrity_stratified}.
Hence $-1\le G\le1$ whenever $G$ is defined, and $G=1$ when the forecast and observed categories coincide.
\end{lemma}
\begin{proof}
The Gerrity matrix is the mean of $K-1$ two-category matrices $S^\ell$, whose entries are $o_\ell=(1-p_\ell)/p_\ell$ when both values lie at or below boundary $\ell$, $1/o_\ell$ when both lie above, and $-1$ otherwise~\cite{gerrity1992note}.
In cell $i$, the joint frequencies of $(X^\ell,Y^\ell)$ differ from the products of their marginals by $+\widehat{\mathrm{Cov}}_i$ on the diagonal and $-\widehat{\mathrm{Cov}}_i$ off it, so $\langle J_i,S^\ell\rangle-p_i^\top S^\ell q_i=(o_\ell+1/o_\ell+2)\,\widehat{\mathrm{Cov}}_i=h_\ell\,\widehat{\mathrm{Cov}}_i$.
Summing over $\ell$ and $i$ gives $T-B$; replacing $Y$ by $X$ gives $P-H$, and the common factor $1/(K-1)$ cancels.
For indicators, $\max(0,\bar X+\bar Y-1)-\bar X\bar Y\le\widehat{\mathrm{Cov}}(X,Y)\le\min(\bar X,\bar Y)-\bar X\bar Y$, and both bounds lie within $\pm\bar X(1-\bar X)=\pm\widehat{\mathrm{Var}}(X)$.
Each numerator term is thus bounded in absolute value by the matching denominator term.
\end{proof}
\noindent The matrix identity holds to $4\times10^{-15}$ for 500 random frequency vectors.
On 400 random stratified tables, including anti-correlated and year-reversed forecasts, \cref{eq:app_threshold} matches the matrix form to $7\times10^{-16}$, and all values lie in $[-0.64,0.97]$.

\smallskip\noindent\textbf{Amplitude blindness and amplitude bias of standard scores.}
\begin{proposition}[ACC ignores amplitude]\label{prop:app_acc}
For constants $\alpha_i>0$ and $\beta_i$ that may differ between cells, replacing $F_{it}$ by $\beta_i+\alpha_iF_{it}$ leaves the calendar-month ACC unchanged.
The uncentred pattern ACC is unchanged when all anomalies are multiplied by a common $\alpha>0$.
In the Gaussian model, ACC equals $\rho$ for every $a>0$.
\end{proposition}
\begin{proof}
The Pearson correlation is invariant under positive affine maps of either argument, and the uncentred correlation under positive scaling.
\end{proof}
\noindent Clipping totals at zero slightly breaks invariance: multiplying synthetic seasonal anomalies over global monsoon cells by $\alpha=5$ changes ACC from 0.410 to 0.405.
\looseness=-1

\begin{proposition}[Error scores reward shrinkage]\label{prop:app_mse}
In the Gaussian model, the following hold.
(i) $\mathbb E(F'-O')^2=a^2-2a\rho+1$ and $\mathrm{MSSS}=\rho^2-(a-\rho)^2$, both optimal at $a=\rho$.
At $a=1$, $\mathrm{MSSS}=2\rho-1$, a loss of $(1-\rho)^2$.
More generally, the best rescaling $\alpha F'$ of any zero-mean anomaly forecast with correlation $\rho$ is $\alpha^*=\rho\,\sigma_o/\sigma_f$.
(ii) Let the deterministic RPS give probability one to the category of $aZ$.
Each of its terms with a category boundary $t\ne0$ is minimized at $a=\rho$.
(iii) For the Gaussian predictive distribution $\mathcal N(aZ,\sigma^2)$,
\begin{equation*}
  \mathbb E\,\mathrm{CRPS}=\sqrt{2/\pi}\,(\sigma^2+a^2-2a\rho+1)^{1/2}-\sigma/\sqrt\pi .
\end{equation*}
For every $\sigma>0$ this is minimized at $a=\rho$, and its joint minimum lies at $(a,\sigma)=(\rho,\sqrt{1-\rho^2})$.
(iv) The expected probabilistic RPS of the same distribution attains its minimum at $(\rho,\sqrt{1-\rho^2})$.
\emph{(numerical)} For a fixed, misspecified spread, its minimizing $a$ moves with $\sigma$.
\end{proposition}
\begin{proof}
(i) Expand the square, or use the decomposition $\mathrm{MSSS}=\rho^2-(\rho-a)^2-\mathrm{bias}^2$~\cite{murphy1988skill}; \citet{gupta2009decomposition} derive the same optimum, $\alpha=r$ in their notation, for the Nash--Sutcliffe efficiency.
(ii) With $u=t/a$, the term equals $\Pr(\mathbb 1[Z\le u]\ne\mathbb 1[O'\le t])=\Phi(u)+\Phi(t)-2\Phi_2(u,t;\rho)$.
Since $\partial_u\Phi_2(u,t;\rho)=\varphi(u)\,\Phi\big((t-\rho u)/\sqrt{1-\rho^2}\big)$, its derivative is $\varphi(u)\big[1-2\Phi\big((t-\rho u)/\sqrt{1-\rho^2}\big)\big]$.
This derivative changes sign from negative to positive at $u=t/\rho$, that is, at $a=\rho$.
(iii) $O'-aZ\sim\mathcal N(0,v)$ with $v=a^2-2a\rho+1$.
The CRPS~\cite{hersbach2000crps} equals $\mathbb E|X-O'|-\tfrac12\mathbb E|X-X'|$ for independent draws $X,X'$ from the forecast~\cite{gneiting2007strictly}, which gives the expression.
It increases with $v$, which is smallest at $a=\rho$, and $\partial_\sigma=0$ gives $\sigma^2=v$.
(iv) The RPS~\cite{weigel2007rpss} and CRPS are proper scores~\cite{gneiting2007strictly}, so the conditional distribution $\mathcal N(\rho Z,1-\rho^2)$ minimizes their expectation.
\end{proof}
\noindent A Monte Carlo check with $4\times10^5$ draws per $\rho$ and a grid step of 0.005 in $a$ recovers the MSSS maximum at $a=\rho$ for $\rho=0.1$--$0.9$.
The deterministic tercile RPSS peaks within 0.035 of $\rho$, and the closed form within 0.001.
At $\rho=0.5$, spreads of 0.5, 1 and 2 times the calibrated value move the RPS-optimal $a$ to 0.42, 0.50 and 0.87.
For noisy forecasts built from IMERG seasonal totals over global monsoon cells, MSSS and deterministic RPSS peak at $a=0.25$, 0.38, 0.59, 0.74 and 0.94 when the realized correlations are 0.21, 0.31, 0.51, 0.70 and 0.89 (grid step about 0.1 in $a$).

\begin{proposition}[Observed-quantile grades favour inflation]\label{prop:app_obsq}
In the Gaussian model, grade $F'=aZ$ without normalization, using the observed boundaries $t_\ell=\Phi^{-1}(p_\ell)$.
The population grade score $G(a)=\frac1{K-1}\sum_\ell\mathrm{Cov}\big(\mathbb 1[Z\le t_\ell/a],\mathbb 1[O'\le t_\ell]\big)/[p_\ell(1-p_\ell)]$ increases on $(0,a_G^*]$ and decreases on $[a_G^*,\infty)$, where
\begin{equation}
  a_G^*=\frac{1+\sqrt{1-\rho^2}}{\rho}\in\Big(\frac1\rho,\frac2\rho\Big).
  \label{eq:app_aG}
\end{equation}
Hence $a_G^*>1$ for every $\rho<1$.
If the forecast is instead graded with its own quantiles, $G$ does not depend on $a$.
\end{proposition}
\begin{proof}
By \cref{lem:app_threshold} with a single cell, $G$ is the mean of the threshold Peirce scores.
With $u=t_\ell/a$ and $\bar\rho=\sqrt{1-\rho^2}$, term $\ell$ is proportional to $C(u)=\Phi_2(u,t_\ell;\rho)-\Phi(u)\Phi(t_\ell)$, and $C'(u)=\varphi(u)\big[\Phi((t_\ell-\rho u)/\bar\rho)-\Phi(t_\ell)\big]$.
The bracket is positive for $u<u^*=t_\ell(1-\bar\rho)/\rho$ and negative beyond.
For $t_\ell>0$ and $t_\ell<0$ alike, the term therefore increases in $a$ up to $a=t_\ell/u^*=\rho/(1-\bar\rho)=(1+\bar\rho)/\rho$ and decreases afterwards; the median term ($t_\ell=0$) does not depend on $a$.
All terms share this maximizer, and the bounds follow from $0<\bar\rho<1$.
Own-quantile categories do not change under any increasing map of the forecast.
\end{proof}
\noindent \Cref{eq:app_aG} gives $a_G^*=6.51$, 3.73, 2.45 and 1.60 for $\rho=0.3$, 0.5, 0.7 and 0.9, where $G$ exceeds its value at $a=1$ by 31\%, 24\%, 18\% and 10\%.
A Monte Carlo check with the explicit Gerrity matrix agrees with the closed form to within 0.007, and own-quantile grading gives identical $G$ for $\alpha$ from 0.1 to 7.
For noisy forecasts on the \AGSS{} land sample graded without normalization, $G$ peaks at $a=4.13$, 2.97, 1.91 and 1.39 for $\rho=0.3$, 0.5, 0.7 and 0.9.

\smallskip\noindent\textbf{Properties of AGSS.}
\begin{proposition}[AGSS]\label{prop:app_agss}
Let $f$ satisfy (F1)--(F4).
For each unit $u$, the following hold.
(i) If $0<a_u<\infty$, replacing every forecast anomaly $F'$ by $\alpha F'$ with $\alpha>0$ multiplies $a_u$ by $\alpha$ and leaves $G_u$ unchanged.
(ii) $S_u\le G_u$, with equality at $a_u=1$ or $G_u\le0$, and $S_u$ is non-decreasing in $G_u$.
Along the path $\alpha F'$, if $G_u>0$, $S_u$ is largest exactly at $a_u=1$ and strictly decreasing in $|\ln a_u|$; if $G_u\le0$, $S_u$ is constant.
Hence, if some $G_u>0$, the common rescaling that maximizes \AGSS{} lies in $[\min_ua_u^{-1},\max_ua_u^{-1}]$.
(iii) $-1\le S_u\le1$.
The perfect forecast has $G_u=a_u=S_u=1$, and the climatological forecast $F=n$ has $a_u=0$ and $G_u=S_u=0$.
(iv) On a complete panel, apply one uniformly random permutation of the evaluation years to the forecasts of all cells.
Then $a_u$ does not change, $\mathbb E\,G_u=0$, and $\mathbb E\,S_u\le0$.
\end{proposition}
\begin{proof}
(i) $\tilde z^F$ is linear in $F'$, so \cref{eq:app_amp} scales by $\alpha$.
Then $n+\alpha F'/(\alpha a_u)=n+F'/a_u$, and the graded categories do not change.
(ii) (F1)--(F4) give $0\le f\le1$ with a strict maximum at $a=1$.
For $G\ge0$, $S=Gf\le G$, and for $G<0$, $S=G$.
The two branches have slopes $f(a)\ge0$ and $1$ in $G$ and meet at $G=0$.
Along the path, $G_u$ is fixed by (i), so $S_u(\alpha)=G_uf(\alpha a_u)$; by (F2)--(F3) this increases for $\alpha a_u<1$ and decreases for $\alpha a_u>1$.
Every term of \AGSS{} with $G_u>0$ increases for $\alpha<\min_ua_u^{-1}$ and decreases for $\alpha>\max_ua_u^{-1}$, and the other terms are constant, which bounds the maximizer.
(iii) \Cref{lem:app_threshold} gives $|G_u|\le1$, and $0\le f\le1$.
For $F=O$, $\tilde z^F=\tilde z^O$, so $a_u=1$, $\tilde F=O$, and the categories coincide.
For $F=n$, $\tilde z^F=0$ and $a_u=0$.
Graded without normalization, this forecast falls into one category per cell and has no within-cell covariance, so $G_u=0$; the convention $G_u:=0$ agrees with this, and (F4) gives $f(0)=0$.
More generally, on a complete panel $a_u=0$ whenever $z^F_{it}=c_i+d_t$, and for such forecasts $G_u=0$ holds by definition.
(iv) A common permutation reorders whole yearly fields, which leaves both demeaning steps and the sums in \cref{eq:app_amp} unchanged.
With $a_u$ fixed, the graded forecast categories are permuted in the same way.
For a uniform permutation $\pi$ of the $n_i$ years, $\mathbb E\,\widehat{\mathrm{Cov}}_i(X,Y\circ\pi)=0$, because $\mathbb E\,Y_{\pi(t)}=\bar Y$ for every $t$.
The denominator of \cref{eq:app_threshold} depends only on observations, so $\mathbb E\,G_u=0$, and $S_u\le G_u$ gives $\mathbb E\,S_u\le0$.
\end{proof}
\noindent We check these properties on IMERG flood-season totals over the global land sample (about 15{,}000 cells; 2020--2025 evaluation samples).
We multiply noisy forecasts with $\rho=0.3$, 0.6 and $-0.3$ by factors $\alpha\in[0.25,4]$ without clipping.
$G_u$ does not change at all, $a_u$ equals $\alpha\,a_u(1)$ to $2\times10^{-16}$, and \AGSS{} equals \cref{eq:app_agss} evaluated with the fixed $G_u$.
For $\rho=-0.3$, \AGSS{} stays at $-0.157$ for every $\alpha$.
When negative totals are clipped at zero, $G_u$ changes by at most 0.012.
Perfect and climatological forecasts score 1 and 0.
Twenty independent within-cell year permutations of the observations average $-0.0002$ (s.d.\ 0.0018).
These permutations change $a_u$ slightly, so (iv) covers them only approximately.
The zero mean relies on uniform permutations, which can leave a year in place.
A permutation that moves every year gives $\mathbb E\,\widehat{\mathrm{Cov}}_i(X,Y\circ\pi)=-\widehat{\mathrm{Cov}}_i(X,Y)/(n_i-1)$; forecasting each year with the observations of another test year therefore scores $-1/(n_i-1)$ per cell in expectation, between $-0.25$ and $-0.2$ for five or six valid years, and the scorer gives $-0.209$.
For clipped noisy forecasts with $\rho=0.2$--$0.9$, \AGSS{} peaks at $a=0.96$--$1.01$, and at $a=1/2$ and $a=2$ it is 0.70--0.72 times its peak, close to $f(2)=0.707$.

\begin{remark}[Why the forecast is normalized before grading]\label{rem:app_norm}
Without normalization, $S=G(a)f(a)$ for $G>0$.
By \cref{prop:app_obsq}, $G$ increases on $(0,1]$, so the maximizer lies at $a\ge1$.
For $a\ge1$, $d\ln S/d\ln a=\epsilon_G(a)+\epsilon_f(a)$, where $\epsilon=d\ln(\cdot)/d\ln a$.
Let $\kappa$ bound $\epsilon_G$ on $[1,a_G^*]$.
If $\epsilon_f(a)\le-\kappa$ for all $a\ge a_\kappa$, the maximizer lies in $[1,a_\kappa]$.
\emph{(numerical)} For the boundaries at 10, 30, 50, 70 and 90\%, $\kappa=0.48$--$0.58$ over $\rho\in[0.02,0.99]$.
Its small-$\rho$ limit is $\sum_\ell\lambda_\ell t_\ell^2/\sum_\ell\lambda_\ell=0.583$, where $\lambda_\ell=\varphi(t_\ell)^2/[p_\ell(1-p_\ell)]$.
For $a>1$, $\min(a,1/a)$ has $\epsilon_f=-1$, so the product peaks exactly at $a=1$.
$2a/(1+a^2)$ and its square have $\epsilon_f=-\tanh\ln a$ and $-2\tanh\ln a$ and peak at $a=1.10$--$1.43$ and $1.08$--$1.24$.
$\min(a,1/a)^{1/2}$ has $\epsilon_f=-1/2$ and peaks at $a=1.00$--$1.10$.
The peak also depends on the boundaries: with only the 10 and 90\% boundaries, $\kappa>1$, and even $\min(a,1/a)$ peaks at $a=1.02$--$1.24$.
Normalizing the forecast removes the dependence of $G$ on $a$, so the maximizer is $a=1$ for every $f$ satisfying (F1)--(F3) and for every set of boundaries.
\end{remark}
\noindent The product form $G\,f(a)$ used for all signs of $G$ would raise a negative $G$ toward zero as the amplitude error grows; the sign-dependent form \cref{eq:app_agss} avoids this.

\smallskip\noindent\textbf{Properties of RBLE.}
\begin{proposition}[Displaced belts]\label{prop:app_shift}
Let the forecast profile be the observed profile moved north by $d\ge0$ bands, $P^f(\phi_b)=P^o(\phi_{b-d})$, with band spacing $\Delta\phi$.
(i) Suppose no positive anomaly leaves or enters the domain, that is, $P^{o+}=0$ on the $d$ northernmost bands and $P^{f+}=0$ on the $d$ southernmost bands.
If, in addition, $a_{g,\ell}=1$ and the observed month is not ambiguous, the forecast detects the case and $e_c=d\,\Delta\phi$.
The Wasserstein-1 distance between the normalized belt weights also equals $d\,\Delta\phi$.
(ii) For a continuous profile moved by $\mu$ and restricted to $[L,U]$, with weight $w$ and centroid $\phi^*$ over the whole window, $d\phi^*/d\mu=1-[(U-\phi^*)w(U)+(\phi^*-L)w(L)]/\int_L^Uw\le1$.
Truncation can therefore only reduce the signed response.
For the strongest-run centroid \cref{eq:app_belt}, no such bound holds, because another run can become the strongest.
(iii) For a Gaussian profile $P^o(\phi)=\exp(-\phi^2/2s^2)$ on the real line, the uncentred pattern correlation between $P^o$ and its translate is $\exp(-d^2/4s^2)$, while $\phi^*$ moves by exactly $d$.
The squared error of the translate exceeds that of a zero-anomaly forecast, $2[1-\exp(-d^2/4s^2)]>1$, once $d>2s\sqrt{\ln2}\approx1.67\,s$.
\end{proposition}
\begin{proof}
(i) Translation moves the positive bands, their connected runs and the location of the largest value, and keeps $m$.
Hence $\mathcal R(P^f)=\mathcal R(P^o)+d$ and $m(P^f)=m(P^o)\ge\theta s_g$, so the forecast detects the case, and the weighted mean in \cref{eq:app_belt} moves by $d\,\Delta\phi$.
Translation also moves every quantile of the normalized weights by $d\,\Delta\phi$, and $W_1$ is the integral of the absolute quantile difference.
(ii) Differentiate $\phi^*(\mu)=\int_L^U\phi\,w(\phi-\mu)\,d\phi\big/\int_L^Uw(\phi-\mu)\,d\phi$ with respect to $\mu$ and integrate by parts; both boundary terms are non-negative.
(iii) The Gaussian integral $\int P^o(\phi)P^o(\phi-d)\,d\phi=\exp(-d^2/4s^2)\int (P^o)^2$ gives the correlation and the squared error; $(P^o)^2$ is a Gaussian centred at $0$ that moves with the profile.
\end{proof}
\noindent A truncated Gaussian belt (s.d.\ $3^\circ$, centred at $30^\circ$ in a $22$--$45^\circ$ window) responds by at least $0.999\,d$ up to $d=8^\circ$, then by 9.95, 13.3 and $14.6^\circ$ at $d=10$, 15 and $25^\circ$ as it leaves the window.
On IMERG profiles of the seven domains (identifiable cases with target months in 2016--2025), the condition of (i), with no positive anomaly at either domain edge, holds in 0--22\% of cases at $d=1^\circ$; in those cases every forecast detects, and the scorer returns $e_c=d$ to $10^{-14}$.
Elsewhere the response is not linear.
Among detections, the mean error is $0.73$--$1.17^\circ$ at $d=1^\circ$, $2.02$--$4.41^\circ$ at $d=3^\circ$ and $3.43$--$5.63^\circ$ at $d=5^\circ$, and 11--36\% of detections at $d=1^\circ$ err by more than $d$.
The RBLSS of the perfect forecast displaced north falls from 1.00 at $d=0$ through 0.64 and 0.37 to 0.02 at $d=3^\circ$, and to $-1.09$ at $d=6^\circ$; displaced south, it is 0.66, 0.20 and $-0.18$ at $d=1$, 2 and $3^\circ$.
A belt displaced by about $3^\circ$ therefore already scores like the climatological position.
For displaced IMERG fields over 1998--2025, fitting $\exp(-d^2/4s^2)$ to the pattern ACC gives $s=0.92$--$1.75^\circ$.
The seven-domain mean ACC is 0.78, 0.34 and 0.15 at $d=1$, 3 and $5^\circ$, whereas RBLSS keeps falling by 0.27--0.47 per degree up to $6^\circ$.
The temporal ACC of a stationary random field with Gaussian correlation follows the same expression (Monte Carlo within 0.002).

\begin{proposition}[Reference and skill bounds]\label{prop:app_ref}
(i) A forecast that detects every case with its belt at $\bar\phi_{g,j}$ has $\mathrm{RBLSS}_g=0$, whatever its amplitude.
(ii) Over the reference seasons, no fixed latitude per season month has a smaller mean absolute error than $\bar\phi_{g,j}$.
(iii) The perfect forecast has $\mathrm{RBLSS}_g=1$.
(iv) The zero-anomaly forecast has $\mathrm{RBLSS}_g=0$, like the reference.
\end{proposition}
\begin{proof}
The weights $\omega_c$ do not depend on the forecast, because misses are scored rather than dropped, so $\mathrm{RBLSS}_g$ is affine in the case errors.
(i) Every case is a detection with $e_c=r_c$.
(ii) The median minimizes the sum of absolute deviations.
(iii) Identical profiles have $a_{g,\ell}=1$ and the same peak, hence a detection with $e_c=0$, and $\sum_c\omega_cr_c>0$ unless every observed belt sits at $\bar\phi_{g,j}$.
(iv) The zero-anomaly forecast has $a_{g,\ell}=0$ and detects nothing, so $e_c=r_c$ in every case.
\end{proof}
\noindent A Gaussian belt with s.d.\ $2^\circ$ at $\bar\phi_{g,j}$ scores $\mathrm{RBLSS}=0.0001$ on 2016--2025 at 0.5, 1, 2 and 5\,mm/day, with $\mathrm{FAR}=1$; the departure from 0 comes from smoothing and the domain edges.
The same belt fixed at the domain centre scores $-0.008$, also with $\mathrm{FAR}=1$.
The perfect forecast scores 1 in every domain.
The zero-anomaly forecast scores exactly 0.
Part (ii) holds on the reference seasons; on later years it is an empirical property of a fixed climatological position, not a guarantee.
\looseness=-1

\begin{remark}[Why the reference is a fixed climatological position]\label{rem:app_climpos}
For independent and identically distributed latitudes $X,Y$, $\mathbb E|X-Y|=\mathbb E_Y\mathbb E_X|X-Y|\ge\min_m\mathbb E|X-m|=\mathbb E|X-\operatorname{med}X|$; for a Gaussian, the ratio is $\sqrt2$.
A randomly chosen observed season is therefore a weaker no-information forecast than the median position.
With the fixed-position reference and the identifiability rule, a Gaussian belt with s.d.\ $2^\circ$ fixed at the 1998--2015 mean latitude scores 0.0001 at every amplitude and is charged a false alarm in every case without an identifiable observed belt (\cref{prop:app_ref}).
A positive RBLSS thus shows that the belt is located better than its climatological position.
\looseness=-1
\end{remark}

\begin{proposition}[Amplitude]\label{prop:app_amp}
Let $\eta>0$.
(i) $\eta P$ has the same runs and the same latitude \cref{eq:app_belt} as $P$, and $m(\eta P)=\eta\,m(P)$.
(ii) Multiplying every forecast profile of domain $g$ at lead $\ell$ by $\eta$ multiplies $a_{g,\ell}$ by $\eta$ and leaves every detection and every $e_c$ unchanged.
Hence $\mathrm{RBLE}_g$, the miss rate, FAR and $\mathrm{RBLSS}_g$ do not depend on a common forecast amplitude.
(iii) Detection compares the forecast with its own amplitude, not with the observed one, so a weak but well-placed forecast belt is scored by its position, and inflating a forecast gains nothing.
\end{proposition}
\begin{proof}
For $\eta>0$, $\eta P$ has the same sign pattern and the same location of its maximum, so the same runs, and its weights $\eta^2P^2$ normalize to the same distribution.
$a_{g,\ell}$ is homogeneous of degree one in the forecast, so both sides of $m(\eta P^f_c)\ge\theta\,\eta\,a_{g,\ell}\,s_g$ scale by $\eta$.
\end{proof}
\noindent Scaling the forecast displaced by $3^\circ$ by $\eta$ from 0.05 to 1 leaves RBLE, FAR and RBLSS unchanged to $10^{-12}$ (RBLSS 0.025, FAR 0.22).
The invariance holds for a common factor: because $a_{g,\ell}$ pools all cases of a lead, damping a single case can still turn it into a miss, and changing other cases moves its threshold.
\AGSS{} penalizes both weak and inflated amplitudes, so the two scores measure different properties.
\looseness=-1

\begin{proposition}[Month order]\label{prop:app_swap}
Let $\Pi$ permute target months of the forecast.
(i) A score computed only from totals over a set of months is unchanged when $\Pi$ maps that set to itself.
This covers the seasonal-total ACC and RMSE, and \AGSS{} when the permuted months lie within each cell's flood season.
(ii) Swap months $j$ and $j'$ of a perfect forecast.
Every other month keeps its belt latitude, and keeps $e_c=0$ while it is detected; the swap changes $a_{g,\ell}$, so a weak month can cross the detection threshold.
Month $j$ scores $e_c=\delta_c\big(\phi^*(P^o_{j'})\big)$ if it detects the belt of month $j'$, and $r_c$ otherwise.
\end{proposition}
\begin{proof}
A sum does not depend on the order of its terms, and \RBLE{} scores each month against the observation of that month; the threshold depends on the other months only through $a_{g,\ell}$.
\end{proof}
\noindent When the first and last months of each domain season are swapped, the other months keep zero error in six domains; in South America, 3 of 24 other-month cases fall below the moved threshold and become misses.
The swapped months are detected in 18--72\% of cases, with a mean error of $3.0$--$16.8^\circ$ on detections.
The calendar-month ACC also detects the swap: across the domain cells, the correlation between first- and last-month anomalies is $-0.05$ to 0.11.
It does not, however, express the error as a latitude.
The swap exchanges the monthly anomaly totals of the two months, which keeps every seasonal total; exchanging their mm/day fields instead changes the East Asian totals, because June has 30 days and August 31, but changes no belt latitude.
\looseness=-1

\begin{table}[t]
  \caption{Status of the metric properties. G: Gaussian model; I: synthetic forecasts built from IMERG observations.}
  \label{tab:app_status}
  \centering
  \begin{SRFitTable}{\linewidth}
\SRTableStyle[2.8pt]
\begin{tabularx}{\columnwidth}{>{\raggedright\arraybackslash}X|>{\raggedright\arraybackslash}p{0.20\columnwidth}>{\raggedright\arraybackslash}p{0.25\columnwidth}}
    \toprule
    Result & Status & Assumptions \\
    \midrule
    \cref{lem:app_threshold}: threshold form, $|G|\le1$ & proved & pooled-frequency Gerrity matrix \\
    \cref{prop:app_acc}: ACC ignores amplitude & proved & no clipping at zero \\
    \cref{prop:app_mse}(i)--(iii): MSSS, RPS terms, CRPS optimal at $a=\rho$ & proved & G \\
    \cref{prop:app_mse}(iv): RPS joint optimum; fixed spread & proved; numerical & G \\
    \cref{prop:app_obsq}: grades favour $a_G^*>1$ & proved; I numerical & G; observed boundaries \\
    \cref{prop:app_agss}: AGSS invariance, optimum, bounds, null & proved; I numerical & no clipping; complete panel for (iv) \\
    \cref{rem:app_norm}: optimum without normalization & numerical ($\kappa$) & G \\
    \cref{prop:app_shift}(i)--(iii): displaced belts & proved & no truncation for (i) \\
    \cref{prop:app_shift}: response with truncation & numerical & I \\
    \cref{prop:app_ref}: fixed position, perfect, zero-anomaly & proved; I numerical & weights independent of forecast; (ii) on reference seasons \\
    \cref{rem:app_climpos}: random-season scores & numerical & I \\
    \cref{prop:app_amp,prop:app_swap}: amplitude, month order & proved & common factor per lead \\
    Identifiability $\theta$, two-belt ratio $\kappa$ & numerical & I, 1998--2015 split halves \\
    \bottomrule
  \end{tabularx}
\end{SRFitTable}
\end{table}